\documentclass[sigconf]{acmart}

\usepackage{graphicx}
\usepackage{booktabs}
\usepackage{adjustbox}
\usepackage{multirow}
\usepackage{fancybox}
\usepackage{tcolorbox}
\usepackage{xcolor}
\usepackage{color, colortbl}
\definecolor{Gray}{gray}{0.9}
\usepackage{caption}

\usepackage{algorithm}
\usepackage{algorithmic}
\usepackage{lscape}

\usepackage{enumitem}

\usepackage{subfigure}
\usepackage{subcaption}

\usepackage{amsmath,amsfonts}
\usepackage{textcomp}

\newtheorem{lemma}{Lemma}
\newtheorem{proposition}{Proposition}
\newtheorem{theorem}{Theorem}
\newtheorem{definition}{Definition}

\AtBeginDocument{%
  }

\setcopyright{acmlicensed}
\copyrightyear{2026}
\acmYear{2026}
\acmDOI{XXXXXXX.XXXXXXX}

\acmConference[WSDM 2026]{The 19th ACM International Conference on Web Search and Data Mining}{February 22--26, 2026}{Boise, Idaho, USA}

\acmISBN{978-1-4503-XXXX-X/18/06}

\begin{document}

\title[GrapHoST]{Does Homophily Help in Robust Test-time Node Classification?}

\author{Yan Jiang}
\email{yan.jiang@uq.edu.au}
\orcid{0009-0008-0641-3259}
\affiliation{%
  \institution{The University of Queensland}
  \city{Brisbane}
  \state{Queensland}
  \country{Australia}
}

\author{Ruihong Qiu}
\email{r.qiu@uq.edu.au}
\affiliation{%
  \institution{The University of Queensland}
  \city{Brisbane}
  \state{Queensland}
  \country{Australia}
}

\author{Zi Huang}
\email{helen.huang@uq.edu.au}
\affiliation{%
  \institution{The University of Queensland}
  \city{Brisbane}
  \state{Queensland}
  \country{Australia}
}

\renewcommand{\shortauthors}{Jiang et al.}

\begin{abstract}

Homophily, the tendency of nodes from the same class to connect, is a fundamental property of real-world graphs, underpinning structural and semantic patterns in domains such as citation networks and social networks. Existing methods exploit homophily through designing homophily-aware GNN architectures or graph structure learning strategies, yet they primarily focus on GNN learning with training graphs. However, in real-world scenarios, test graphs often suffer from data quality issues and distribution shifts, such as domain shifts across users from different regions in social networks and temporal evolution shifts in citation network graphs collected over varying time periods. These factors significantly compromise the pre-trained model's robustness, resulting in degraded test-time performance. With empirical observations and theoretical analysis, we reveal that transforming the test graph structure by increasing homophily in homophilic graphs or decreasing it in heterophilic graphs can significantly improve the robustness and performance of pre-trained GNNs on node classifications, without requiring model training or update. Motivated by these insights, a novel test-time graph structural transformation method grounded in homophily, named GrapHoST, is proposed. Specifically, a homophily predictor is developed to discriminate test edges, facilitating adaptive test-time graph structural transformation by the confidence of predicted homophily scores. Extensive experiments on nine benchmark datasets under a range of test-time data quality issues demonstrate that GrapHoST consistently achieves state-of-the-art performance, with improvements of up to 10.92\%. Our code has been released at \href{https://github.com/YanJiangJerry/GrapHoST}{\textcolor{purple}{https://github.com/YanJiangJerry/GrapHoST}}.
\end{abstract}



\maketitle
\section{Introduction}

In real-world graph structural data, homophily plays a crucial role in characterising structural and semantic patterns across graphs from various domains. The concept of \textbf{homophily}, originating from sociology, refers to the tendency of nodes from the same class (e.g., friends with similar characteristics in social networks) to form connections~\cite{mcpherson2001birds}. In contrast, \textbf{heterophily}, or low homophily, describes the tendency of nodes from different classes to be connected~\cite{lozares2014homophily, heterophily, Geom-GCN}, as in a movie collaboration network, where actors associated with their distinct genres appear together in the same movies~\cite{Geom-GCN}.

\begin{figure}[!t]
\centering
\includegraphics[width=0.9\linewidth]{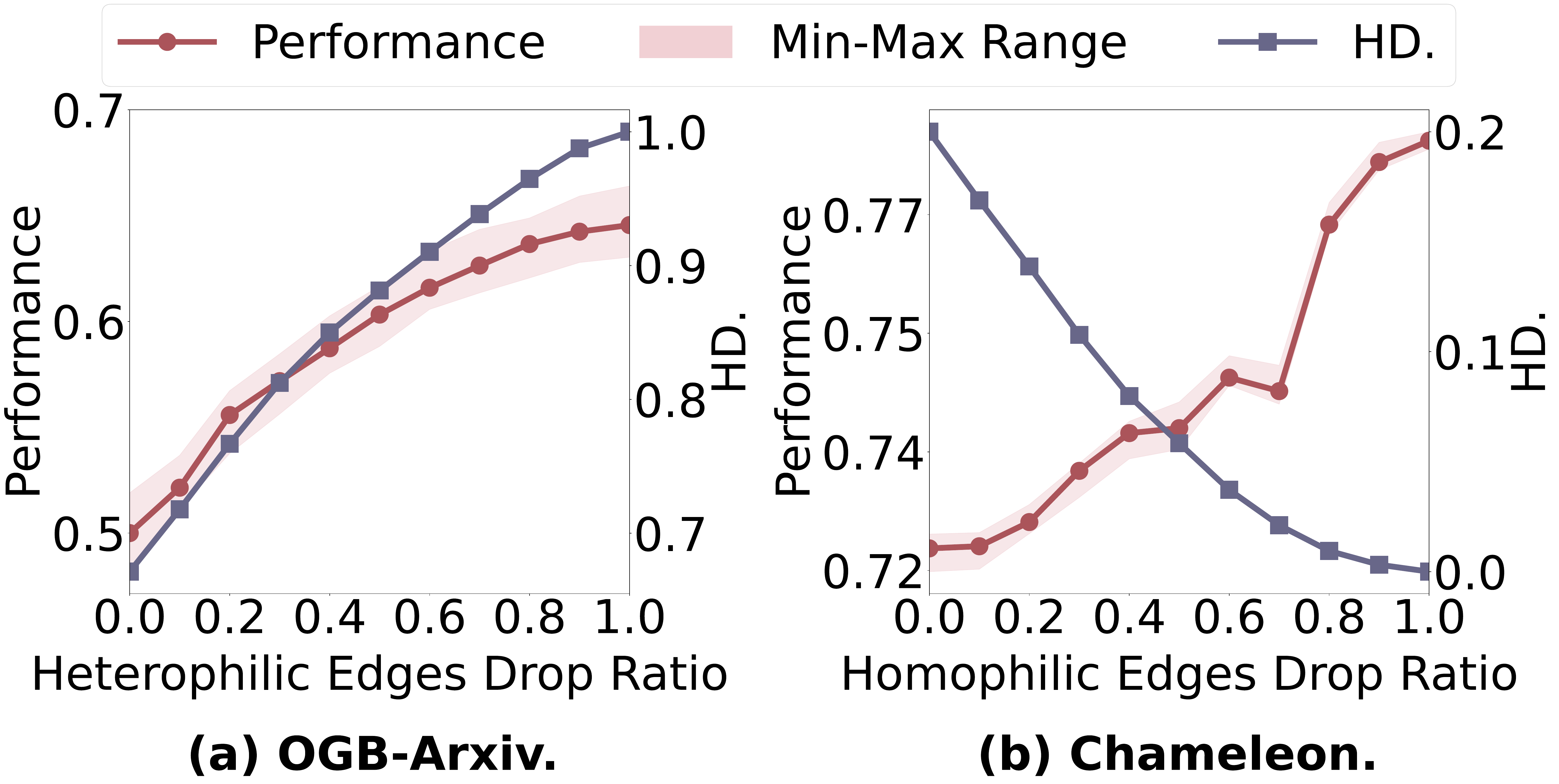} 
\vspace{-0.4cm}
\caption{Empirical observations of test node classification performance for a fixed pre-trained GNN under data quality issues. HD.\ is the edge homophily degree. (a) For a homophilic test graph with temporal shift (Arxiv, HD.=67.13\%), increasing the test graph’s homophily degree improves the pre-trained GNN's performance. (b) For a heterophilic test graph with node attribute shift (Chameleon, HD.=22.52\%), decreasing the homophily degree improves performance.
}
\vspace{-0.8cm}
\label{fig:motivation}
\end{figure}

Most existing methods aim to train or update a GNN to cope with the homophily patterns in the training graph. A mainstream of methods has focused on developing homophily-aware GNNs that incorporate homophilic and heterophilic information from static training graphs for model learning~\cite{GREET, GOAL, h2gcn, linkx, bo2021beyond}. Additionally, graph structure learning~\cite{Opengsl} aims to train GNNs with a modified training graph based on homophily~\cite{HOLE, huang2025multiplex}. More recent test-time training methods start to employ model re-training based on the static test graph to capture homophily patterns~\cite{Matcha, HomoTTT, HGDA, ASSESS}.

Although these methods have made progress in leveraging homophily information during GNN training~\cite{zhu2020beyond, ma2021homophily, ND, HSBM, ACM, LHS, HGDA}, they lack guarantees that the trained classifier will perform robustly across diverse real-world testing scenarios, particularly when test graphs suffer from data quality issues, which result in suboptimal model test-time performance~\cite{EERM, GTRANS}. For instance, such challenges often arise from domain shifts, such as social network graphs collected from users in different regions~\cite{rozemberczki2021multi}, or citation networks constructed from distinct time periods~\cite{EERM}. To this end, this paper aims to address these challenges by posing the following question:
\vspace{-0.2cm}
\begin{tcolorbox}[
  colframe={rgb,1:red,0.75;green,0.60;blue,0.45},  
  colback={rgb,1:red,0.98;green,0.96;blue,0.93},   
  coltitle=black,
  boxsep=1pt, left=6pt, right=6pt, top=10pt, bottom=10pt
]
\vspace{-0.2cm}
\textbf{In testing scenarios upon various data quality issues, how do the homophily-related properties in test graphs influence the pre-trained GNNs?}
\vspace{-0.2cm}
\end{tcolorbox}
\vspace{-0.2cm}

To answer this question, an empirical study of test graph homophily property is conducted in Figure~\ref{fig:motivation}, and a theoretical analysis is later detailed in Section~\ref{sec:theory}. In Figure~\ref{fig:motivation} (a), with \textbf{a fixed GNN pre-trained on a homophilic training graph, increasing the edge homophily degree} (mathematical definition in Section~\ref{sec:notation}) of the test graph by removing heterophilic edges can lead to significant improvement of the model testing performance, such as for the Arxiv graph~\cite{OGB}. Similarly, in Figure~\ref{fig:motivation} (b), with \textbf{a fixed GNN pre-trained on a heterophilic training graph, decreasing the edge homophily degree} of the test graph by removing homophilic edges can also improve the model testing performance, such as for the Chameleon graph~\cite{Geom-GCN}. These findings are further verified by \textbf{a novel perspective of using Contextual Stochastic Block Model (CSBM)~\cite{CSBM, GPR, HSBM, ma2021homophily} to theoretically analyse the change in graph instead of change in model} in Section~\ref{sec:theory}. However, due to the absence of ground-truth labels in test graphs, distinguishing between homophilic and heterophilic edges is still challenging. Although some methods augment the test graph using differentiable techniques~\cite{GTRANS} or node addition~\cite{GP}, they can hardly exploit the specific homophily information in the test graph.

To effectively leverage such homophily-based properties in test graphs without label access, this paper proposes a novel test-time graph structural transformation framework based on homophily, named GrapHoST, to effectively improve the quality of the test graphs by (\romannumeral 1) increasing the homophily degree of homophilic scenarios or (\romannumeral 2) decreasing the homophily degree of heterophilic scenarios by transforming the test graph structure in a fine-grained edge level. Specifically, a homophily predictor is designed to learn valuable edge homophily patterns from training graphs and provide homophily-based predictions for the test graph. The test graph structure is then transformed, enabling the pre-trained GNN classifier to achieve improved performance. In contrast to previous graph structure learning methods, \textbf{GrapHoST is a model-agnostic and data-centric method, which does not require updating the GNN classifier}, which makes GrapHoST a plug-and-play module into existing graph learning frameworks. Our contributions are:
\begin{itemize}[left=0em, itemsep=0pt, topsep=0pt, parsep=0pt, partopsep=0pt]
\item The first study of the relationship between test graph homophily properties and the test-time performance of pre-trained GNNs, supported by empirical observation and theoretical analysis.
\item A novel GrapHoST to improve fixed GNNs' performance on test graphs with data quality issues, from a data-centric perspective.
\item Extensive experiments on nine benchmark datasets with various data quality issues demonstrate the robustness and state-of-the-art performance of GrapHoST, with improvements up to 10.92\%.
\end{itemize}

\begin{figure*}[!t]
\centering
\includegraphics[width=0.87\linewidth]{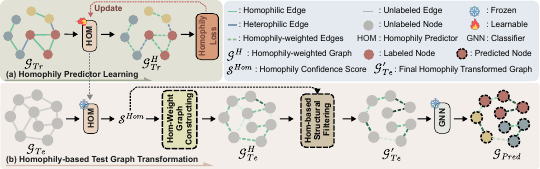} 
\vspace{-0.2cm}
\caption{Overall framework for homophilic graphs. (a) A homophily predictor learns the homophily-based properties to classify edges as homophilic or heterophilic. (b) In the testing scenario, the learned homophily predictor estimates homophily confidence scores. Based on these scores, a homophily-weighted graph is constructed, followed by a confidence-aware homophily-based edge filtering process. The homophily-based transformed test graph is then passed to a fixed GNN classifier for enhanced test-time performance. A similar pipeline can be applied to heterophilic graphs, with predicted homophilic edges filtered out.}
\vspace{-0.2cm}
\label{fig:method}
\end{figure*}

\vspace{-0.2cm}
\section{Preliminary}
\label{sec:notation}
\textbf{Graph notations}. A graph is defined as $\mathcal{G} = (\mathcal{V}, \mathcal{E})$, where $\mathcal{V} = \{1, 2, \ldots, n\}$ is the set of nodes and $\mathcal{E} = \{e_{ij}\}$ is the set of edges. 
Each node is associated with a label $y$ within $c$ unique classes, and $\mathcal{Y}$ is the set of labels of all nodes in the graph. The edge homophily degree is defined as $HD. = \big|\{e_{ij} \mid e_{ij}\in \mathcal{E}, y_{i} = y_{j}\}\big|/|\mathcal{E}|$.

\subsection{Problem Definition}
\label{sec:problem}
\begin{definition}
\textbf{(Test-time Graph Structural Transformation)}. Given a fixed pre-trained $\text{GNN}_{\boldsymbol{\beta}^*}$ with parameter $\boldsymbol{\beta}$ optimised over the training graph $\mathcal{G}_{Tr} = (\mathcal{V}_{Tr}, \mathcal{E}_{Tr})$ and $\mathcal{Y}_{Tr}$, this task aims to learn a graph structural transformation $f_\theta(\cdot)$ that modifies the structure of the test graph $\mathcal{G}_{Te} = (\mathcal{V}_{Te}, \mathcal{E}_{Te})$ into $\mathcal{G}_{Te}'=(\mathcal{V}_{Te},f(\mathcal{G}_{Te}))$, such that node classifications of $\mathcal{V}_{Te}$ by $\text{GNN}_{\boldsymbol{\beta}^*}$ is improved:
\begin{align} 
\arg \min_{\theta} &\mathcal{L} \left(\text{GNN}_{\boldsymbol{\beta}^*}(\mathcal{V}_\text{Te}, f_\theta(\mathcal{G}_\text{Te})), \mathcal{Y_\text{Te}}\right) \text{ s.t. } f_\theta(\mathcal{G}_\text{Te}) \in \mathcal{P}(\mathcal{E}_\text{Te}), \notag \\
\text{with } \boldsymbol{\beta}^* &= \arg \min_{\boldsymbol{\beta}} \mathcal{L} \left(\text{GNN}_{\boldsymbol{\beta}} (\mathcal{V}_\text{Tr},\mathcal{E}_\text{Tr}), \mathcal{Y_\text{Tr}}\right),
\end{align}
where $\mathcal{L}$ is the loss function to assess downstream performance; $\mathcal{P}(\mathcal{E}_{Te})$ denotes the possible edge spaces of the test graph $\mathcal{G}_{Te}$.
\end{definition}

\section{Theoretical Analysis}
\label{sec:theory}
\textbf{CSBM}. The theoretical analysis will be based on Contextual Stochastic Block Model (CSBM)~\cite{CSBM, GPR, HSBM, ma2021homophily}, a graph generative model for random graphs. We will start a CSBM with two classes, $ c_1 $ and $ c_2 $ (\textbf{multi-class} extension is natural and provided in \textcolor{purple}{Supp.~\ref{sec:proof-multi} in Supplementary Material}). Edges are generated according to an intra-class probability $ p $ and an inter-class probability $ q $. In a homophilic graph, the probability of homophilic edges exceeds that of heterophilic edges ($ p > q $). For each node $ i $, its feature vector $ \mathbf{x_i} \in \mathbb{R}^l $ is sampled from a Gaussian distribution $ \mathbf{x_i} \sim \mathcal{N}(\boldsymbol{\mu_k}, \mathbf{I}) $, where $ \boldsymbol{\mu_k} \in \mathbb{R}^l $ represents the mean feature vector for class $ c_k $ in dimension $l$, with $ k \in \{1,2\} $, and $ \boldsymbol{\mu_1} \neq \boldsymbol{\mu_2} $. Therefore, a graph generated by CSBM can be defined as $ \mathcal{G} \sim \text{CSBM}(\boldsymbol{\mu_1}, \boldsymbol{\mu_2}, p, q) $.

Unlike previous work that adopts CSBM to analyse different GNNs' behaviour on the same graph~\cite{ND, CSBM, ma2021homophily}, this work leverage CSBM \textbf{for a fixed GNN on a test graph and the structurally transformed test graph}, thus bridging the gap between the intrinsic homophily-based properties of test graphs and the GNN test-time performance.

To begin with, the embedding of node $ i $ after GNN encoding operations is denoted as $ \mathbf{h}_i $. For the original test graph $\mathcal{G}_{Te}$, the expectations of node embeddings of both classes can be denoted as $\mathbb{E}_{c_1}[\mathbf{h}_i]$ and $\mathbb{E}_{c_2}[\mathbf{h}_i]$. Similarly for the test graph after structural transformation $\mathcal{G}'_{Te}$, embeddings are $\mathbb{E}_{c_1}[\mathbf{h}'_i]$ and $\mathbb{E}_{c_2}[\mathbf{h}'_i]$.

\begin{tcolorbox}[
colframe=gray!60, colback=gray!10, coltitle=black, boxsep=0pt, 
left=2pt, right=2pt, top=2pt, bottom=2pt]
\begin{lemma}
\label{lemma:boundary}
\textbf{(Midpoint of embeddings)}. Vectors $ (\mathbb{E}_{c_1}[\mathbf{h}'_i], \mathbb{E}_{c_2}[\mathbf{h}'_i]) $ and $ (\mathbb{E}_{c_1}[\mathbf{h}_i], \mathbb{E}_{c_2}[\mathbf{h}_i]) $ share a same midpoint $\mathbf{m}$, and $ \mathbf{m} = \frac{\boldsymbol{\mu_1} + \boldsymbol{\mu_2}}{2} $. 
\end{lemma}
\end{tcolorbox}

\begin{tcolorbox}[colframe=gray!60, colback=gray!10, coltitle=black, boxsep=0pt, left=2pt, right=2pt, top=2pt, bottom=2pt]
\begin{lemma}
\label{lemma:boundary_2}
\textbf{(Embedding direction)}. The differences $ \mathbb{E}_{c_1}[\mathbf{h}'_i] - \mathbb{E}_{c_2}[\mathbf{h}'_i] $ and $ \mathbb{E}_{c_1}[\mathbf{h}_i] - \mathbb{E}_{c_2}[\mathbf{h}_i] $ are in a same direction $ \mathbf{o}$, and $ \mathbf{o} = \frac{\boldsymbol{\mu_1} - \boldsymbol{\mu_2}}{\|\boldsymbol{\mu_1} - \boldsymbol{\mu_2}\|_2} $. 
\end{lemma}
\end{tcolorbox}

The proof is given in {\textcolor{purple}{Supp.~\ref{sec:proof-lemma}}}, with detailed mathematical definitions of $\mathbf{h}$ and $\mathbb{E}[\mathbf{h}]$ in {\textcolor{purple}{Supp.~\ref{sec:embedding}}}. Upon embeddings $\mathbf{h}$, an optimal decision boundary for a binary classifier can be obtained:

\begin{tcolorbox}[colframe=gray!60, colback=gray!10, coltitle=black, boxsep=0pt, left=2pt, right=2pt, top=2pt, bottom=2pt]
\begin{proposition}
\label{prop:db}
   (\textbf{Optimal Linear Classifier Boundary}). Assuming the number of nodes between classes is balanced, the hyperplane orthogonal to $ \mathbf{o} $ and passing through $ \mathbf{m} $ defines the optimal decision boundary $\mathcal{B}$ of the linear classifier with embeddings $\mathbf{h}$ as follows:
\begin{equation}
\mathcal{B} = \{\mathbf{h} \mid \mathbf{o}^\top \mathbf{h} - \mathbf{o}^\top \mathbf{m} = 0\}.
\end{equation}
\end{proposition}
\end{tcolorbox}
\setcounter{theorem}{0}

Intuitively, the optimal classifier (decision boundary) $\mathcal{B}$ should pass through the midpoint $\mathbf{m}$ between the embeddings $\mathbf{h}$ of the two classes, thereby maximising the classification capability for both classes. A larger separation between the class embeddings and the boundary $\mathcal{B}$ typically indicates improved node classification performance at test time. Considering a homophilic test graph $\mathcal{G}^{Hom}_{Te}$ and its homophily-based transformed test graph $\mathcal{G}'^{Hom}_{Te}$, with a fixed classifier, the following theorems can be derived:
\vspace{-0.1cm}
\begin{tcolorbox}[
  colframe={rgb,1:red,0.75;green,0.60;blue,0.45},  
  colback={rgb,1:red,0.98;green,0.96;blue,0.93}, 
  coltitle=black,
  boxsep=0pt,
  left=2pt,
  right=2pt,
  top=2pt,
  bottom=2pt
]
\begin{theorem}
\label{thm:homo-imp}
    (\textbf{Node Classification on Homophilic Test Graph under a Fixed Classifier}). For node $i$ in a homophilic test graph $\mathcal{G}^{Hom}_{Te} \sim \text{CSBM}(\boldsymbol{\mu_1}, \boldsymbol{\mu_2}, p, q), \text{ where } (p > q)$, with a fixed classifier defined by the decision boundary $\mathcal{B}$ as in Proposition~\ref{prop:db}, if $\mathcal{G}^{Hom}_{Te}$ is transformed into $\mathcal{G'}^{Hom}_{Te} \sim \text{CSBM}(\boldsymbol{\mu_1}, \boldsymbol{\mu_2}, p', q')$ with a higher homophily degree $(p' > p) \text{ and } (q' < q)$, the misclassification probability of $\mathcal{B}$ on $\mathcal{G}'^{Hom}_{Te}$ is lower than $\mathcal{G}^{Hom}_{Te}$. 
\end{theorem}
\end{tcolorbox}

\begin{tcolorbox}[
  colframe={rgb,1:red,0.75;green,0.60;blue,0.45},  
  colback={rgb,1:red,0.98;green,0.96;blue,0.93}, 
  coltitle=black,
  boxsep=0pt,
  left=2pt,
  right=2pt,
  top=2pt,
  bottom=2pt
]
\begin{theorem}
\label{thm:heter-imp}
(\textbf{Node Classification on Heterophilic Test Graph under a Fixed Classifier}). For node $i$ in a heterophilic test graph $\mathcal{G}^{Het}_{Te} \sim \text{CSBM}(\boldsymbol{\mu_1}, \boldsymbol{\mu_2}, p, q), \text{ where } (p < q)$, with a fixed classifier defined by the decision boundary $\mathcal{B}$ as in Proposition~\ref{prop:db}, if $\mathcal{G}^{Het}_{Te}$ is transformed into $\mathcal{G'}^{Het}_{Te} \sim \text{CSBM}(\boldsymbol{\mu_1}, \boldsymbol{\mu_2}, p', q')$ with a lower homophily degree $(p' < p) \text{ and } (q' > q)$, the misclassification probability of $\mathcal{B}$ on $\mathcal{G}'^{Het}_{Te}$ is lower than $\mathcal{G}^{Het}_{Te}$. 
\end{theorem}
\end{tcolorbox}
\vspace{-0.1cm}

The proof is given in {\textcolor{purple}{Supp.~\ref{sec:proof-binary} and~\ref{sec:proof-binary-heter}}}. Empirical visualisations can be found in Figure~\ref{fig:umap}. Intuitively, \textbf{refining the test graph structure only based on homophily properties can directly reduce the misclassification probability under a fixed pre-trained GNN}. There is no hard restriction on the requirement of different types of test-time data quality issues, offering general guidance for improving GNN test-time performance to transform the test graph based on a data-centric perspective on the homophily change. Extensions to \textbf{(1) imbalanced degree} ({\textcolor{purple}{Supp.~\ref{sec:relax-degree}}}), \textbf{(2) imbalanced class} ({\textcolor{purple}{Supp.~\ref{sec:relax-boundary}}}) and \textbf{(3) multi-class} ({\textcolor{purple}{Supp.~\ref{sec:proof-multi}}}) are natural.

\section{Methodology}
\label{sec:method}
To validate the theoretical analysis, this section will present the GrapHoST framework, with illustration in Figure~\ref{fig:method} and Algorithm~\ref{alg:training}. The main objectives are to address the challenges in real-world test \textbf{without touching the base classifier}: (1) how to \textbf{estimate the homophily of edges in test graphs without label access}; and (2) how to \textbf{perform structural transformation on test graphs} based on estimated homophily score with high confidence.

\subsection{Fixed Pre-trained GNN Classifier}
For the node classification tasks, a fixed pre-trained GNN model is employed to encode the nodes into high-dimensional embeddings:
\begin{equation}
\label{eq:def}
\mathbf{Z} = \text{GNN}_{\boldsymbol{\beta}^*}(\mathcal{V},\mathcal{E}),
\end{equation}
where $\mathbf{Z} \in \mathbb{R}^{n \times c}$ is the logit matrix of all $n$ nodes, $c$ is the class number, and $\boldsymbol{\beta}^*$ is the optimised parameter of the fixed GNN classifier.

\subsection{Homophily Predictor Learning}
\label{sec:training}
To adjust the homophily degree of test graphs, an intuitive strategy is to remove heterophilic edges from homophilic graphs and vice versa. However, the absence of ground-truth labels in the test graphs makes it difficult to distinguish the test edges.
To address this challenge, a novel homophily predictor is developed to learn homophily-based properties on the training graph
$\mathcal{G}_{Tr} = (\mathcal{V}_{Tr}, \mathcal{E}_{Tr})$, where the homophilic edges $\mathcal{E}^{Hom}$ are labelled as the positive class and the heterophilic edges $\mathcal{E}^{Het}$ as the negative class. The homophily predictor $\text{HOM}_{\boldsymbol{\theta}}$ can be constructed by a GNN backbone.
\begin{align}
\mathbf{Z}^{Hom} &= \text{HOM}_{\boldsymbol{\theta}}(\mathcal{V}_{Tr},\mathcal{E}_{Tr}),
\end{align}
where $\mathbf{Z}^{Hom}\in\mathbb{R}^{n\times l}$ is the output node embeddings matrix of all $n$ nodes with dimension $l$, and $\mathbf{z}_{i}$ is the embedding of node $i$, encoded by the homophily predictor $\text{HOM}_{\boldsymbol{\theta}}$ with learnable parameter $\boldsymbol{\theta}$. The homophily prediction for each edge can then be obtained as:
\begin{align}
\label{eq:homo_score}
\hat{y}_{ij} &= \sigma(\cos(\mathbf{z}_{i}, \mathbf{z}_{j})), \quad \text{s.t. } e_{ij} \in \mathcal{E}_{Tr},
\end{align}
where $\sigma$ is the sigmoid function, and $\hat{y}_{ij}\in\mathbb{R}$ is the predicted homophily score, representing the probability of the homophily class for edge $e_{ij}$. Note that during the learning of the homophily predictor, only training labels are utilised, thereby facilitating an inductive training process without requiring access to test nodes.

Real-world graphs exhibit diverse homophilic and heterophilic edge patterns, where the number of edges in the two classes is often imbalanced~\cite{Opengsl}. To address edge class imbalance issues during homophily predictor training, a Weighted Binary Cross-Entropy (WBCE) loss is designed to assign a greater weight to the minority edge class. For instance, in the Amazon-Photo dataset, the average edge homophily is 95.46\%, implying that heterophilic edges constitute the minority and are assigned a higher weight based on edge number. The homophily loss is defined as:
\begin{equation}
\label{eq:loss}
\mathcal{L}^{Hom}_{WBCE} = - \sum\nolimits_{i=1}^{n} \Big[ \alpha y_i \log(\hat{y}_i) + (1 - \alpha) (1 - y_i) \log(1 - \hat{y}_i) \Big],
\end{equation}
where $ \alpha = \frac{|\mathcal{E}^{Het}|}{|\mathcal{E}^{Het}| + |\mathcal{E}^{Hom}|} $. The edge label $ y $ is set to 1 if the edge is the homophilic edge $\mathcal{E}^{Hom}$ that connects nodes of the same class, and 0 otherwise. The effectiveness of the homophily predictor trained with homophily loss is analysed in Section~\ref{sec:loss}.

\subsection{Homophily-based Test-time Graph Structural Transformation}
During test time, the homophily confidence score $\mathbf{s}^{Hom}$ for each test edge is computed by the homophily predictor as defined in Eq.~\eqref{eq:homo_score}, defined as $\mathbf{s}^{Hom} = \left[\hat{y}_{ij}\right]_{e_{ij} \in \mathcal{E}_{\mathrm{Te}}}$
Intuitively, this homophily confidence score indicates the predicted probability of homophily for each test edge, ranging from 0 to 1, reflecting the predictor’s confidence in classifying edges as homophilic. Correspondingly, the heterophilic confidence score for each test edge is defined as $\mathbf{s}^{Het} = \mathbf{1} - \mathbf{s}^{Hom}$, which represents the predicted probability of an edge being heterophilic. These homophily-based confidence scores serve as the foundation for enhancing the test graph structure.

\subsubsection{\textbf{Homophily-weighted Test Graph Construction}}
After obtaining homophily confidence scores $\mathbf{s}^{Hom}$ in homophilic graphs and heterophily confidence scores $\mathbf{s}^{Het}$ in heterophilic graphs, our objective is to effectively construct a homophily-weighted structure for the test graphs. Therefore, GrapHoST first proposes a weighted homophily-based test graph, leveraging the homophily-based confidence scores. Specifically, the edge weight is defined as:
\begin{equation}
\begin{aligned}
\label{eq:reweight_edges}
w_{ij} = 
\begin{cases} 
\mathbf{s}^{Hom}(e_{ij}), & \forall e_{ij} \in \mathcal{E}_{Te} \quad \text{in Hom. case}, \\ 
\mathbf{s}^{Het}(e_{ij}), & \forall e_{ij} \in \mathcal{E}_{Te} \quad \text{in Het. case}.
\end{cases}
\end{aligned}
\end{equation}
Compared to the original test graph, the homophily-weighted test graph, defined as $\mathcal{G}_{Te}^{H} = (\mathcal{V}_{Te}, \mathcal{E}_{Te}, \mathcal{W})$, with the set of edge weights $\mathcal{W}$, directly leverages homophily-based properties for an enhanced test graph structure. For example, in homophilic graphs such as Cora, homophilic edges receive higher weights than predicted heterophilic edges, which contain noisy and harmful structural patterns that can hinder the pre-trained classifier's test-time performance. Therefore, the homophily-weighted test graph highlights the important structural patterns in the test graphs, resulting in superior test-time performance, as analysed in Section~\ref{sec:effectiveness}.

\subsubsection{\textbf{Homophily-based Test Graph Structural Filtering}}
Motivated by Theorems~\ref{thm:homo-imp} and~\ref{thm:heter-imp}, homophilic edges in homophilic test graphs are beneficial to the test-time performance of the pre-trained model, whereas their counterpart edges are harmful. To safely transform the test graph structure, GrapHoST uses a confidence-aware edge filtering strategy that prunes only the most confidently predicted harmful edges from the test graphs. Specifically, the Top-$\delta$ ratio of confident heterophilic edges in homophilic test graphs and the Top-$\delta$ ratio of confident homophilic edges in heterophilic test graphs are filtered out by the confidence scores:
\begin{equation}
\begin{aligned}
\label{eq:filtering}
\mathcal{E}_{Filter} = 
\begin{cases} 
\{ e \in \mathcal{E}_{Te} \mid \mathbf{s}^{Het}(e) \geq \delta \} & \text{in Hom. case}, \\
\{ e \in \mathcal{E}_{Te} \mid \mathbf{s}^{Hom}(e) \geq \delta \} & \text{in Het. case},
\end{cases}
\end{aligned}
\end{equation}
where $\delta$ denotes the filtering ratio in the range of (0, 1), corresponding to the top $\delta$ fraction of predicted homophily or heterophily scores. A default choice of 0.3 for $\delta$ can yield satisfactory performance. Further optimal choice can be determined based on the validation data with experiments on hyperparameter sensitivity in Section~\ref{sec:hyper}. A possible automatic design of $\delta$ can be related to the predictive confidence, which will be left to future exploration.

The edges in $\mathcal{E}_{Filter}$ are then filtered from the test graph, yielding $\mathcal{E}'_{Te} = \mathcal{E}_{Te} \setminus \mathcal{E}_{Filter}$, where $\mathcal{E}'_{Te}$ denotes the test graph structure after filtering. As a result, the original test graph $\mathcal{G}_{Te} = (\mathcal{V}_{Te}, \mathcal{E}_{Te})$ is transformed into a filtered homophily-weighted test graph $\mathcal{G}'_{Te} = (\mathcal{V}_{Te}, \mathcal{E}'_{Te}, \mathcal{W})$. Notably, the total test-time process is label-free and requires no modification to the pre-trained GNN classifier.

\begin{algorithm}[!t]
\caption{GrapHoST Framework}
\label{alg:training}
\begin{algorithmic}[1]
\REQUIRE 
Fixed pre-trained classifier $\text{GNN}_{\boldsymbol{\beta}^*}$, learnable homophily predictor $\text{HOM}_{\boldsymbol{\theta}}$, training graph $\mathcal{G}_{Tr} = (\mathcal{V}_{Tr}, \mathcal{E}_{Tr})$, and test graph  $\mathcal{G}_{Te} = (\mathcal{V}_{Te}, \mathcal{E}_{Te})$
\ENSURE
Prediction $\hat{\mathbf{Y}}_{Te}$ and optimised $\boldsymbol{\theta}^*$
\STATE \textcolor{blue}{/*Homophily predictor learning*/}
\WHILE{not converged}
    \STATE Optimise $\text{HOM}_{\boldsymbol{\theta}}$ with $\mathcal{L}^{Hom}_{WBCE}$ on $\mathcal{G}_{Tr}$ to obtain $\boldsymbol{\theta}^*$ (Eq.~\eqref{eq:loss})
\ENDWHILE
\STATE \textcolor{blue}{/*Test graph structural transformation*/} 
\STATE Obtain edge homophily confidence scores $\mathbf{s}^{Hom}$ with $\text{HOM}_{\boldsymbol{\theta}^*}$ and $\mathcal{G}_{Te}$
\STATE Obtain homophily-reweighted test graph $\mathcal{G}_{Te}^{H}$ (Eq.~\eqref{eq:reweight_edges})
\STATE Obtain structurally filtered test graph $\mathcal{G}'_{Te}$ (Eq.~\eqref{eq:filtering})
\STATE \textcolor{blue}{/*Inference on transformed test graph with pre-trained GNN*/}
\STATE Predict $\hat{\mathbf{Y}}_{Te}$ with $\text{GNN}_{\boldsymbol{\beta}^*}$ and $\mathcal{G}'_{Te}$ (Eq.~\eqref{eq:inference2})
\RETURN $\hat{\mathbf{Y}}_{Te}$ and $\boldsymbol{\theta}^*$
\end{algorithmic}
\end{algorithm}

\subsection{Model Inference}
After obtaining the homophily-structure-enhanced test graphs, the fixed pre-trained GNN classifier takes the enhanced test graph structure along with the original node features as input to predict node classes. Notably, the GNN operation defined in Eq.~\eqref{eq:def} is extended to incorporate the homophily-based edge weight set $\mathcal{W}$ without modifying the model parameters, as follows:
\begin{align}
\label{eq:inference}
\mathbf{Z}_{Te}' &= \text{GNN}_{\boldsymbol{\beta}^*}(\mathcal{V}_{Te},\mathcal{E}’_{Te}, \mathcal{W}).
\end{align}

The GNN classifier performs message passing on the homophily-based transformed test graph and encodes nodes into embeddings:
\begin{equation}
\mathbf{h}_i^{(l+1)} = \sigma \left( \sum\nolimits_{j \in \mathcal{N}(i)} w_{ij} \mathbf{W}^{(l)} \mathbf{h}_j^{(l)} + \mathbf{b}^{(l)} \right),
\end{equation}
where $\mathbf{h}_i^{(0)}=\mathbf{x}_i$, $w_{ij}$ is the weight of edge $e_{ij}$ from the edge weight set $\mathcal{W}$, $l$ denotes the layer number. The matrix $\mathbf{W}^{(l)}$ and $\mathbf{b}^{(l)}$ are the weight and bias of the fixed GNN Classifier in layer $l$, and $\mathcal{N}(i)$ denotes the neighbours of node $i$. The embeddings from the last layer are the logit matrix $\mathbf{Z}_{Te}'$, which can be fed into a softmax function to obtain the predicted class probability $\hat{\mathbf{Y}}_{Te}$ for all nodes:
\begin{equation}
\begin{aligned}
\label{eq:inference2}
\hat{\mathbf{Y}}_{Te} &= \text{softmax}(\mathbf{Z}_{Te}').
\end{aligned}
\end{equation}

\section{Discussions}
This section includes discussions on GrapHoST, with additional content in \textcolor{purple}{Supp.~\ref{app:discuss}}.
\paragraph{\textbf{Mixed homophilic and heterophilic patterns in real-world graphs}}
Real-world graphs typically exhibit diverse homophily patterns~\cite{patterns}. The datasets adopted for GrapHoST also demonstrate mixed homophily–heterophily structures. As summarised in Table~\ref{tab:summary}, Amazon-Photo has the highest edge homophily degree (95.46\%), followed by Elliptic, Arxiv, Cora, and Twitch. \textbf{To evaluate the effectiveness of GrapHoST on graphs with more complex and mixed homophily patterns}, the FB100 and Twitch datasets are adopted. Such graph datasets have an average edge homophily degree of 53.23\% and 57.22\%, indicating the substantial coexistence of both homophily and heterophily patterns~\cite{lim2021new}. As in Section~\ref{sec:exp}, GrapHoST consistently shows effectiveness on such graphs with mixed patterns. More detailed analysis on the mixed and complex homophily patterns in real-world datasets are provided in {\textcolor{purple}{Supp.~\ref{sec:homo-shift}}.}

\paragraph{\textbf{How GrapHoST differs from graph structure learning methods on homophily}}
Existing graph structure learning methods that target homophily, such as GOAL~\cite{GOAL} and HoLe~\cite{HOLE}, focus on exploiting structural patterns in static graphs to facilitate improved graph structure learning during GNN training, without considering the importance and potential of homophily-based structural patterns present in test graphs~\cite{zhu2020beyond, HOLE, ma2021homophily, ND, HSBM, ACM, GREET, LHS, GOAL}. In contrast, GrapHoST specifically targets test time, investigating the role of homophily-based test structural patterns in the performance of a fixed pre-trained GNN, which is fundamentally different from all existing graph structure learning methods.

\paragraph{\textbf{Why GrapHoST works}}
The effectiveness of GrapHoST can be attributed to the following factors: (\romannumeral 1) Theoretical motivation and analysis provide a guarantee for improved GNN performance during test-time by adjusting the degree of edge homophily of the test graphs. (\romannumeral 2) The homophily predictor, trained with a weighted binary cross-entropy loss, enhances its effectiveness on test graphs with quality issues. (\romannumeral 3) Homophily-weighted test graph construction and confidence-aware edge filtering transform the test graph based on the confidence of the homophily scores, mitigating the impact of potential misclassified edges. The combination of these mechanisms ensures the effectiveness of GrapHoST.

\section{Experiment}
\label{sec:exp}

In experiments, the following research questions (RQs) are studied:
\begin{itemize}[left=0em, itemsep=0pt, topsep=0pt, parsep=0pt, partopsep=0pt]
\item \textbf{RQ1:} How does GrapHoST perform compared to baselines?
\item \textbf{RQ2:} How does each module affect GrapHoST?
\item \textbf{RQ3:} How robust is GrapHoST against extreme noise?
\item \textbf{RQ4:} How does GrapHoST affect test graphs?
\item \textbf{RQ5:} How efficient is GrapHoST compared to existing methods?
\item \textbf{RQ6:} How sensitive is GrapHoST to hyperparameters?
\end{itemize}

\begin{table}[!t]
\centering
\caption{Dataset statistics for homophilic (Hom.) and heterophilic (Het.) graphs. $HD.$ denotes edge homophily degree.}
\label{tab:summary}
\resizebox{\linewidth}{!}{
\begin{tabular}{c|lcccccc}
\toprule
&\textbf{Dataset} & \textbf{\#Nodes} & \textbf{\#Edges} & \textbf{\#Classes} & \textbf{Metric} & \textbf{$HD.$\ (\%)} \\ 
\midrule
\multirow{6}{*}{\rotatebox{90}{Hom.}}&Cora           & 2,703         & 5,278              & 10    & Accuracy  & 67.30 \\
&Photo      & 7,650         & 119,081            & 10    & Accuracy  & 95.46 \\
&Twitch       & 1,912–9,498   & 31,299–153,138     & 2     & F1 Score  & 57.22 \\ 
&FB100          & 769–41,536    & 16,656–1,590,655   & 2     & Accuracy  & 53.23 \\ 
&Elliptic       & 203,769       & 234,355            & 2     & F1 Score  & 69.17 \\ 
&Arxiv      & 169,343       & 1,166,243          & 40    & Accuracy  & 67.13 \\ 
\midrule
\multirow{3}{*}{\rotatebox{90}{Het.}}&Actor          & 7,600         & 30,019             & 5     & Accuracy  & 22.43 \\ 
&Chameleon      & 2,277         & 36,101             & 5     & Accuracy  & 22.52 \\ 
&Squirrel       & 5,201         & 217,073            & 5     & Accuracy  & 22.01 \\ 
\bottomrule
\end{tabular}
}
\vspace{-0.5cm}
\end{table}

\begin{table*}[!t]
    \centering
    \caption{Overall performance with GCN. OOM indicates out of memory on a 32GB V100 GPU. The {\color{purple}\textbf{bold}} denotes the best method, and the {\color{teal}\color{teal}\underline{underline}} is the best baseline. Note that the recent TTT method Matcha~\cite{Matcha} is shown in Table~\ref{tab:overall_2} but not here since it is designed with GPR~\cite{GPR} but not easily extended to other GNNs. Results for more GNN backbones are also in Table~\ref{tab:overall_2}. Results for the homophily predictor implemented by more homophilic or heterophilic GNN backbones, such as H2GCN and FAGCN, are in Table~\ref{tab:backbone}. Average performance and standard deviation are presented in percentage (\%).
    }

    \resizebox{\linewidth}{!}{
        \begin{tabular}{c|c|cccccc|ccc|c}
            \toprule
            &\textbf{Method} & \textbf{Photo} & \textbf{Cora} & \textbf{FB100} & \textbf{Elliptic} & \textbf{Arxiv} & \textbf{Twitch} & \textbf{Actor} & \textbf{Chameleon} & \textbf{Squirrel} & \textbf{Rank} \\
         
            \midrule
            \midrule
            \multirow{3}{*}{\rotatebox{90}{SL}} & ERM (GCN) & 93.02$\pm$1.15 & 92.58$\pm$1.49 & 54.16$\pm$0.92 & 52.35$\pm$6.50 & 50.14$\pm$4.68 & 43.85$\pm$3.48 & 59.14$\pm$0.76 & 72.31$\pm$0.77 & 54.66$\pm$0.95 & 5.33\\
            &DropEdge & 86.28$\pm$4.88 & 79.83$\pm$2.03 & 53.67$\pm$0.88 & 55.96$\pm$6.06 & 49.98$\pm$3.66 & \color{teal}\color{teal}\underline{48.93$\pm$1.70} & 60.16$\pm$0.81 & \color{teal}\color{teal}\underline{73.36$\pm$0.63} & 56.97$\pm$1.05 & 6.11\\
            &EERM & 90.29$\pm$1.34 & 84.76$\pm$2.05 & 54.14$\pm$0.34 & 50.88$\pm$1.97 & OOM & 45.44$\pm$6.47 & \color{teal}\color{teal}\underline{60.46$\pm$0.29} & 73.15$\pm$0.72 & \color{teal}\color{teal}\underline{61.82$\pm$0.31} & 5.50\\
            \midrule
            \multirow{2}{*}{\rotatebox{90}{TTT}}&Tent & 93.23$\pm$1.06 & 92.69$\pm$1.47 & \color{teal}\color{teal}\underline{54.20$\pm$0.96} & 47.15$\pm$1.12 & \color{teal}\color{teal}\underline{50.30$\pm$2.32} & 43.61$\pm$3.54 & 59.13$\pm$0.76 & 72.31$\pm$0.77 & 54.66$\pm$0.95 & 5.67\\
            &FTTT & 91.56$\pm$1.33 & 88.80$\pm$1.12 & 54.16$\pm$0.92 & 54.30$\pm$4.03 & 43.85$\pm$3.02 & 41.49$\pm$3.32 & 59.15$\pm$0.76 & 72.30$\pm$0.75 & 54.72$\pm$0.94 & 6.44\\
            \midrule
            \multirow{3}{*}{\rotatebox{90}{TTGT}}&GraphPatcher & 92.81$\pm$0.56 & 86.75$\pm$1.67 & OOM & \color{teal}\color{teal}\underline{57.99$\pm$6.00} & 45.41$\pm$3.13 & 39.69$\pm$4.81 & 58.77$\pm$1.62 & 71.10$\pm$0.94 & 53.27$\pm$1.26 & 7.38\\
            &GTrans & \color{teal}\color{teal}\underline{93.51$\pm$1.05} & \color{teal}\color{teal}\underline{95.47$\pm$0.67} & 54.10$\pm$0.85 & 56.52$\pm$4.89 & 49.92$\pm$4.68 & 41.39$\pm$1.60 & 59.17$\pm$0.75 & 72.35$\pm$0.82 & 55.06$\pm$0.90 & \color{teal}\color{teal}\underline{5.11}\\

            \cmidrule{2-12}
            \rowcolor{gray!30}
            \cellcolor{white}&\textbf{GrapHoST}  & \color{purple}\textbf{95.94$\pm$0.85} & \color{purple}\textbf{96.12$\pm$0.47} & \color{purple}\textbf{54.68$\pm$0.76} & \color{purple}\textbf{64.32$\pm$3.80} & \color{purple}\textbf{54.78$\pm$2.83} & \color{purple}\textbf{53.52$\pm$1.49} & \color{purple}\textbf{64.07$\pm$0.55} & \color{purple}\textbf{81.17$\pm$0.48} & \color{purple}\textbf{61.86$\pm$0.49} & \color{purple}\textbf{1.22}\\
            
            \bottomrule
        \end{tabular}
    }
    \label{tab:overall_1}
\end{table*}

\textbf{Datasets}. Following prior work~\cite{GTRANS, GP}, experiments are carried out on nine benchmark datasets with dataset statistics shown in Table~\ref{tab:summary}, including six homophilic and three heterophilic graphs with three key data quality issues explored in their test graphs: (1) \textbf{synthetic node attribute shift} in Cora~\cite{yang2016revisiting}, Amazon-Photo~\cite{shchur2018pitfalls} (Photo), Actor, Chameleon, and Squirrel~\cite{Geom-GCN}. (2) \textbf{cross-domain with attribute and structure shift} in Twitch-E~\cite{rozemberczki2021multi} (Twitch) and FB-100~\cite{lim2021new} (FB100), (3) \textbf{temporal evolution with attribute and structure shift} in Elliptic~\cite{Finance} and OGB-Arxiv~\cite{OGB} (Arxiv). Node attribute shift is introduced by adding synthetic noise to node features while keeping the graph structure unchanged, following EERM~\cite{EERM} and GTrans~\cite{GTRANS}. Cross-domain and temporal shifts result from splitting nodes by domain or timestamp~\cite{EERM, GTRANS}. This work strictly follows \textbf{inductive data split} used prior work. For the heterophilic datasets, experiments on the filtered clean datasets following~\cite{heter} with different split settings are available in the {\textcolor{purple}{Supp.~\ref{sec:filter}.}}

\textbf{Baselines}. 8 state-of-the-art baseline methods are compared:
\begin{itemize}[left=0em, itemsep=0pt, topsep=0pt, parsep=0pt, partopsep=0pt]
    \item \textbf{Supervised learning methods (SL)}:
    \begin{itemize}[left=0em, itemsep=0pt, topsep=0pt, parsep=0pt, partopsep=0pt]
        \item \textbf{ERM}: Empirical Risk Minimisation refers to the standard training approach for GNNs as a baseline for comparison~\cite{EERM}.
        \item \textbf{DropEdge}: A method that randomly removes edges during GNN training to improve model performance~\cite{DropEdge}.
        \item \textbf{EERM}: Explore-to-Extrapolate Risk Minimisation to enhance GNNs generalisation during training~\cite{EERM}.
    \end{itemize}

    \item \textbf{Test-time training methods (TTT)}, which re-train the GNN classifier on test graphs while fixing the input test graphs:
    \begin{itemize}[left=0em, itemsep=0pt, topsep=0pt, parsep=0pt, partopsep=0pt]
        \item \textbf{Tent}: A test-time adaptation method that fine-tunes the model's batch normalisation parameters~\cite{Tent}.
        \item \textbf{FTTT}: Fully Test-Time Training, a method proposed by HomoTTT to retrain the GNNs during inference~\cite{HomoTTT}.
        \item \textbf{Matcha}: A test-time training method that adjusts the hop-aggregation parameters in GNNs~\cite{Matcha}.
    \end{itemize}

    \item \textbf{Test-time graph transformation methods (TTGT)}, which fix the pre-trained GNN classifier and modify the test graphs only:
    \begin{itemize}[left=0em, itemsep=0pt, topsep=0pt, parsep=0pt, partopsep=0pt]
        \item \textbf{GTrans}: A method that transforms graphs at test-time to improve fixed pre-trained GNN performance~\cite{GTRANS}.
        \item \textbf{GraphPatcher}: A method that augments the test graph by training an additional model to generate nodes~\cite{GP}.
    \end{itemize}
\end{itemize}

\paragraph{\textbf{Metrics}}
\label{sec:metric}
F1-macro is used to evaluate Elliptic and Twitch due to class imbalance following existing work~\cite{GTRANS}. The mean accuracy is evaluated for all other datasets following existing work~\cite{EERM, GTRANS, GP}. The \textbf{edge homophily degree ($HD.$)} used in this paper is defined as the ratio of edges connecting nodes of the same class~\cite{Opengsl, log, ND, zhao2024all}.

\paragraph{\textbf{Implementation}}
\label{sec:implement}
The edge filtering ratio $\delta$ is chosen from 0 to 0.9 in increments of 0.1, and a default of 0.3 can yield satisfactory results. GNNs are trained with a hidden size of 32 and the number of layers searched from 2 to 5 following baselines~\cite{GTRANS}. Adam optimiser~\cite{ADAM} is adopted with learning rates searching within \{1e-2, 1e-3, 1e-4, 1e-5\}. The default homophily predictor and GNN classifier is GCN. For the homophily predictor, the hidden size and number of layers match the GNN classifier. All experiments are randomly repeated 10 times, with the mean and standard deviation reported. More details on the implementation can be found in the reproducibility statement in {\textcolor{purple}{Supp.~\ref{sec:reproduce}.}}

\subsection{Overall Performance}

The overall performance is evaluated on nine benchmark datasets. To ensure consistency, Table~\ref{tab:overall_1} reports all results based on a fixed GCN classifier on undirected graphs. Results on other GNN classifier backbones are presented in Table~\ref{tab:overall_2}. In addition, Table~\ref{tab:backbone} shows different backbones for the homophily predictor.

According to Table~\ref{tab:overall_1}, for homophilic datasets, it is evident that \textbf{(1) test-time methods, such as GTrans, achieve relatively higher scores than training-time methods, such as EERM, demonstrating the superiority of test-time approaches.} Furthermore, among test-time methods, data-centric methods, such as GTrans and GrapHoST, score better than model-centric methods like Tent and FTTT, indicating that \textbf{(2) directly improving the test graph enhances GNN performance more effectively than retraining the model.} Within data-centric test-time methods, GrapHoST consistently outperforms GTrans and GraphPatcher, demonstrating \textbf{(3) the superiority of a test-time graph transformation method by homophily-based properties.} For heterophilic datasets, GrapHoST outperforms all baselines by a large margin, showing that improving the quality of heterophilic test graphs by homophily-based properties also boosts GNN test-time performance. To be noticed, for large graphs, such as FB100 and Arxiv, existing methods such as GraphPatcher and EERM encounter out-of-memory issues, whereas \textbf{GrapHoST exhibits great scalability across large-scale graphs.}

\begin{table}[!t]
    \centering
    \caption{Performance on other GNNs. Test-time methods are shown. Matcha~\cite{Matcha} is only compatible with GPR.}

    \resizebox{\linewidth}{!}{
        \begin{tabular}{c|c|ccccc}
            \toprule
            & \textbf{Method} & \textbf{Elliptic} & \textbf{Arxiv} & \textbf{Twitch} & \textbf{Actor} \\
            \midrule
            \midrule
            \multirow{4}{*}{\rotatebox{90}{GAT}} & ERM (GAT) & 63.45$\pm$1.90 & \color{teal}\underline{52.90$\pm$2.46} & 45.81$\pm$7.93 & 56.36$\pm$1.59 \\
            & GraphPatcher & 58.81$\pm$3.92 & 48.69$\pm$3.78 & 45.03$\pm$8.71 & 45.38$\pm$0.51 \\
            & GTrans & \color{teal}\underline{66.03$\pm$1.92} & 45.05$\pm$6.22 & \color{teal}\underline{48.32$\pm$5.98} & \color{teal}\underline{56.37$\pm$1.60} \\
            
            \cmidrule{2-6}
            \rowcolor{gray!30}
            \cellcolor{white}& \textbf{GrapHoST}  & \color{purple}\textbf{66.57$\pm$2.93} & \color{purple}\textbf{53.03$\pm$2.41} & \color{purple}\textbf{55.27$\pm$3.41} & \color{purple}\textbf{61.46$\pm$1.18} \\
            
            \midrule
            \midrule
            \multirow{4}{*}{\rotatebox{90}{SAGE}} & ERM (SAGE) & 67.37$\pm$3.20 & 52.55$\pm$2.31 & 38.94$\pm$3.62 & \color{teal}\underline{65.16$\pm$0.81} \\
            & GraphPatcher & \color{teal}\underline{68.25$\pm$3.25} & 51.06$\pm$2.40 & 37.05$\pm$4.19 & 57.55$\pm$1.20 \\
            & GTrans & 68.01$\pm$2.71 & \color{teal}\underline{53.85$\pm$1.51} & \color{teal}\underline{42.74$\pm$3.52} & 62.90 $\pm$0.88 \\
            \cmidrule{2-6}
            \rowcolor{gray!30}
            \cellcolor{white}& \textbf{GrapHoST}  & \color{purple}\textbf{71.38$\pm$2.42} & \color{purple}\textbf{53.89$\pm$1.51} & \color{purple}\textbf{51.44$\pm$2.42} & \color{purple}\textbf{67.35$\pm$0.62} \\
            
            \midrule
            \midrule
            \multirow{6}{*}{\rotatebox{90}{GPR}} & ERM (GPR) & 64.22$\pm$4.40 & 56.69$\pm$0.74 & 29.00$\pm$7.32 & 62.10$\pm$0.97 \\
            & GraphPatcher & 62.72$\pm$4.28 & 56.67$\pm$1.04 & 34.72$\pm$8.98 & 62.63$\pm$0.59 \\
            & GTrans & \color{teal}\underline{70.44$\pm$2.60} & 56.33$\pm$0.73 & 32.36$\pm$7.06 & 61.35$\pm$1.02 \\
            \cmidrule{2-6}
            & Matcha & 67.25$\pm$3.09 & \color{teal}\underline{56.75}$\pm$1.24 & \color{teal}\underline{35.07}$\pm$5.98 & \color{teal}\underline{62.90}$\pm$1.34 \\
            \cmidrule{2-6}
            \rowcolor{gray!30}
            \cellcolor{white}& \textbf{GrapHoST}  & \color{purple}\textbf{70.56$\pm$2.64} & \color{purple}\textbf{57.57$\pm$0.71} & \color{purple}\textbf{37.18$\pm$7.90} & \color{purple}\textbf{64.13$\pm$0.65} \\
            \bottomrule
        \end{tabular}
    }
    \label{tab:overall_2}
\end{table}

\begin{table}[!t]
    \centering
    \caption{Performance of different homophily predictors.
    }
    \vspace{-0.2cm}
    \resizebox{\linewidth}{!}{
        \begin{tabular}{c|l|cccc}
            \toprule
            &\textbf{Method} & \textbf{Elliptic} & \textbf{Arxiv} & \textbf{Twitch} & \textbf{Actor} \\
            \midrule
            \midrule
            &ERM (GCN) & 52.35$\pm$6.50 & 50.14$\pm$4.16 & 43.85$\pm$3.48 & 59.14$\pm$0.76 \\
            \midrule
            &HOM-MLP & 62.45$\pm$4.93 & 53.23$\pm$3.37 & 53.45$\pm$2.79 & 63.49$\pm$0.67 \\
            \midrule
            \multirow{4}{*}{\rotatebox{90}{Hom GNN}} &HOM-GCN & \color{purple}\textbf{64.32$\pm$3.80} & \color{purple}\textbf{54.78$\pm$2.83} & 53.52$\pm$1.49 & \color{teal}\underline{64.07$\pm$0.55} \\
            &HOM-GAT & 62.24$\pm$4.31 & \color{teal}\underline{53.29$\pm$3.26} & \color{purple}\textbf{53.99$\pm$0.56} & 63.39$\pm$0.68 \\
            &HOM-SAGE & \color{teal}\underline{63.61$\pm$4.32} & 52.62$\pm$3.41 & \color{teal}\underline{53.90$\pm$0.43} & 62.62$\pm$0.54 \\
            &HOM-GPR & 62.55$\pm$4.28 & 51.28$\pm$3.72 & 52.79$\pm$1.34 & 62.33$\pm$0.54 \\
            \midrule
            \multirow{3}{*}{\rotatebox{90}{Het GNN}} &HOM-H2GCN & 62.04$\pm$4.31 & 51.38$\pm$4.32 & 48.36$\pm$1.10 & 63.92$\pm$0.66 \\
            &HOM-LINKX & 62.66$\pm$4.32 & 53.21$\pm$3.36 & 51.72$\pm$2.07 & \color{purple}\textbf{64.28$\pm$0.65} \\
            &HOM-FAGCN & 63.52$\pm$2.98 & 53.25$\pm$3.71 & 52.24$\pm$1.52 & 64.05$\pm$1.16 \\
            \bottomrule
        \end{tabular}
    }
    \label{tab:backbone}
    \vspace{-0.5cm}
\end{table}

\textbf{GrapHoST with different GNN classifier backbones}. To validate the effectiveness of GrapHoST, experiments are conducted on more backbone GNN classifiers, including GraphSAGE~\cite{sage}, GAT~\cite{gat}, and GPR~\cite{GPR}. As shown in Table~\ref{tab:overall_2}, GrapHoST can effectively improve base classifiers with different GNN backbones, as well as the GCN from the overall table. Note that the latest test-time training method, Matcha~\cite{Matcha} is only compatible with specific types of GNNs, such as GPR~\cite{GPR}, which is also shown in Table~\ref{tab:overall_2}.

\textbf{Homophily predictor with different backbones}. This experiment in Table~\ref{tab:backbone} evaluates with the performance of the homophily predictor (HOM) using (1) MLPs; (2) GNNs for homophilic graphs (Hom GNN: GAT~\cite{gat}, GraphSAGE~\cite{sage}, and GPR~\cite{GPR}); and (3) GNNs tailored for heterophilic settings (Het GNN: H2GCN~\cite{h2gcn}, FAGCN~\cite{bo2021beyond} and LINKX~\cite{linkx}). Overall, \textbf{the homophily predictor with various architectures consistently improves the base GCN classifier}, while cross-architecture performance over different backbones are robust across datasets.

\subsection{Ablation Study}
\begin{figure}[!h]
\centering
    \begin{minipage}[t]{0.9\linewidth}
        \centering
        \includegraphics[width=\linewidth]{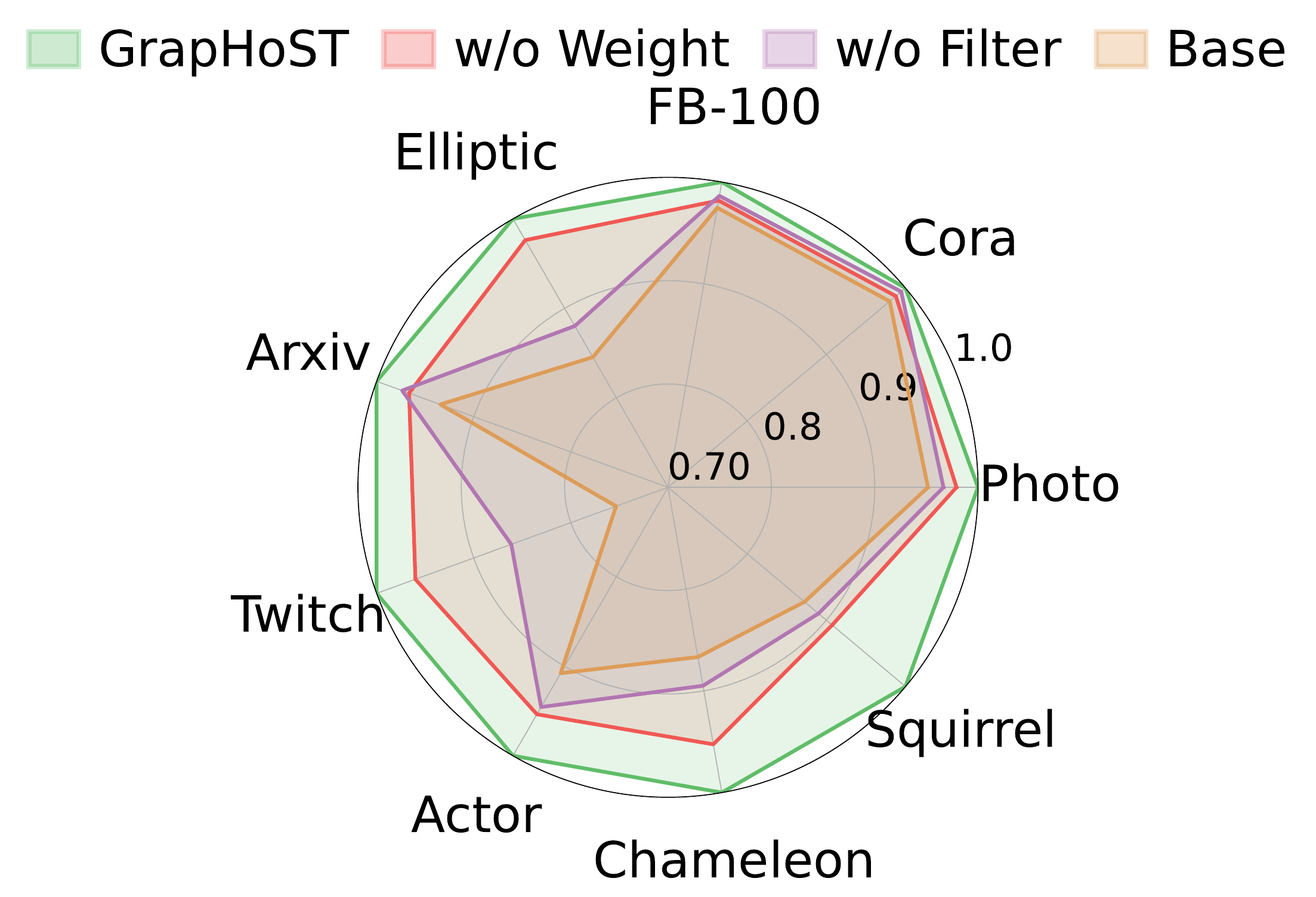}
    \end{minipage}
\caption{
Ablation study of GrapHoST. ``Base" refers to the performance of the base GCN by ERM.
}
\label{fig:ablation}
\end{figure}

An ablation study is conducted to validate the effectiveness of each module in GrapHoST. Two components of GrapHoST are evaluated: (\romannumeral 1) \textit{w/o weight}, removing the homophily-weighted graph construction part; and (\romannumeral 2) \textit{w/o filter}, removing the edge filtering part. Base refers to the original GNN performance by ERM. According to Figure~\ref{fig:ablation}, the full GrapHoST method outperforms all variants. Both edge-filtering and homophily-weight component methods contribute to the overall performance. Results verify the \textbf{effectiveness of each component in GrapHoST}.

\subsection{Robustness against Structural Noise}
This experiment evaluates the robustness of the proposed homophily predictor in various noisy settings and under extreme test-time structural noise. Different ratios of random edge addition and deletion are applied to the test graph. For instance, \textit{GrapHoST + 50\%} indicates that 25\% of the original test edges are randomly removed, and an equal number of randomly sampled non-existent edges are added as noise to the original test graphs. As shown in Table~\ref{tab:robust}, \textbf{even with significant structural noise, GrapHoST} with the homophily predictions can consistently \textbf{outperform base GCN under original test graphs} without such manually added structural noise across all datasets.

\begin{table}[!h]
    \centering
    \caption{Robustness under different levels of structural noise.}
    \resizebox{1.0\linewidth}{!}{
        \begin{tabular}{l|cccc}
            \toprule
            & \textbf{Photo} & \textbf{Cora} & \textbf{Elliptic} & \textbf{Arxiv} \\
            \midrule \midrule
            ERM (GCN + 0\%) & $93.02 \pm 1.15$ & $92.58 \pm 1.49$ & $52.35 \pm 6.50$ & $50.14 \pm 4.68$ \\
            \midrule
            \rowcolor{gray!30}GrapHoST + 0\% & $\mathbf{95.94 \pm 0.85}$ & $\mathbf{96.12 \pm 0.47}$ & $\mathbf{64.32 \pm 3.80}$ & $\mathbf{54.78 \pm 2.83}$ \\
            GrapHoST + 10\% & $95.66 \pm 0.89$ & $95.47 \pm 0.82$ & $62.29 \pm 4.31$ & $54.37 \pm 2.75$ \\
            GrapHoST + 30\% & $94.90 \pm 0.88$ & $94.69 \pm 1.10$ & $62.27 \pm 4.27$ & $53.28 \pm 2.60$ \\
            GrapHoST + 50\% & $93.79 \pm 0.90$ & $93.90 \pm 1.24$ & $62.02 \pm 4.30$ & $51.38 \pm 2.37$ \\
            \bottomrule
        \end{tabular}
    }
    \label{tab:robust}
\end{table}

\subsection{How GrapHoST Impacts Test Graphs}
\label{sec:effectiveness}
\subsubsection{\textbf{Plug-and-Play of GrapHoST on Different GNN Training Strategies}}
As a plug-and-play approach, GrapHoST can be seamlessly integrated into GNNs trained by other methods. As demonstrated in Table~\ref{tab:plug}, incorporating GrapHoST into static GNNs trained with DropEdge and EERM results in substantial improvements in GNN test-time performance, \textbf{highlighting its effectiveness in enhancing other methods}. A random edge-dropping method that ignores homophily-based properties is further implemented as a baseline for comparison with GrapHoST, with the number of pruned edges kept consistent. Results show that random edge-dropping degrades the performance of the fixed GNN at test time, whereas GrapHoST improves it. This is because \textbf{random dropping disrupts critical structural patterns, while homophily-based edge filtering preserves and highlights informative homophily-based connectivity patterns in test graphs}.

\begin{table}[!h]
    \centering
    \caption{Comparison between GrapHoST and random edge dropping on static GNNs.}
    \resizebox{\linewidth}{!}{
        \begin{tabular}{l|cccc}
            \toprule
            \textbf{Method} & \textbf{Photo} & \textbf{Cora} & \textbf{Elliptic} & \textbf{Arxiv} \\
            \midrule
            \midrule
            ERM (GCN) & 93.02$\pm$1.15 & 92.58$\pm$1.49 & 52.35$\pm$6.50 & 50.14$\pm$4.16 \\
            +Random & 93.32$\pm$2.25 & 92.24$\pm$1.79 & 50.44$\pm$7.72 & 46.54$\pm$4.94 \\
            \rowcolor{gray!30}\textbf{+GrapHoST} & \textbf{95.94$\pm$0.85} & \textbf{96.12$\pm$0.47} & \textbf{64.32$\pm$3.80} & \textbf{54.78$\pm$2.83} \\
           
            \midrule
            DropEdge & 86.28$\pm$4.88 & 79.83$\pm$2.03 & 55.96$\pm$6.06 & 49.98$\pm$3.66 \\
            +Random & 84.49$\pm$6.47 & 77.62$\pm$2.74 & 52.54$\pm$6.19 & 45.43$\pm$4.18 \\
            \rowcolor{gray!30}\textbf{+GrapHoST} & \textbf{87.26$\pm$6.39} & \textbf{89.97$\pm$0.47} & \textbf{65.73$\pm$2.57} & \textbf{50.43$\pm$4.73} \\
            
            \midrule
            EERM & 90.29$\pm$1.34 & 84.76$\pm$2.05 & 50.88$\pm$1.97 & OOM \\
            +Random & 89.02$\pm$2.21 & 81.94$\pm$2.49 & 49.88$\pm$3.29 & --- \\
            \rowcolor{gray!30}\textbf{+GrapHoST} & \textbf{95.05$\pm$0.86} & \textbf{85.43$\pm$1.92} & \textbf{53.36$\pm$5.09} & --- \\
            
            \bottomrule
        \end{tabular}
    }\label{tab:plug}
\end{table}

\subsubsection{\textbf{Effectiveness of Edge Filtering in Homophily Degree Change}}
\begin{figure}[!h]
\centering
    \begin{minipage}[t]{\linewidth}
        \centering
        \includegraphics[width=\linewidth]{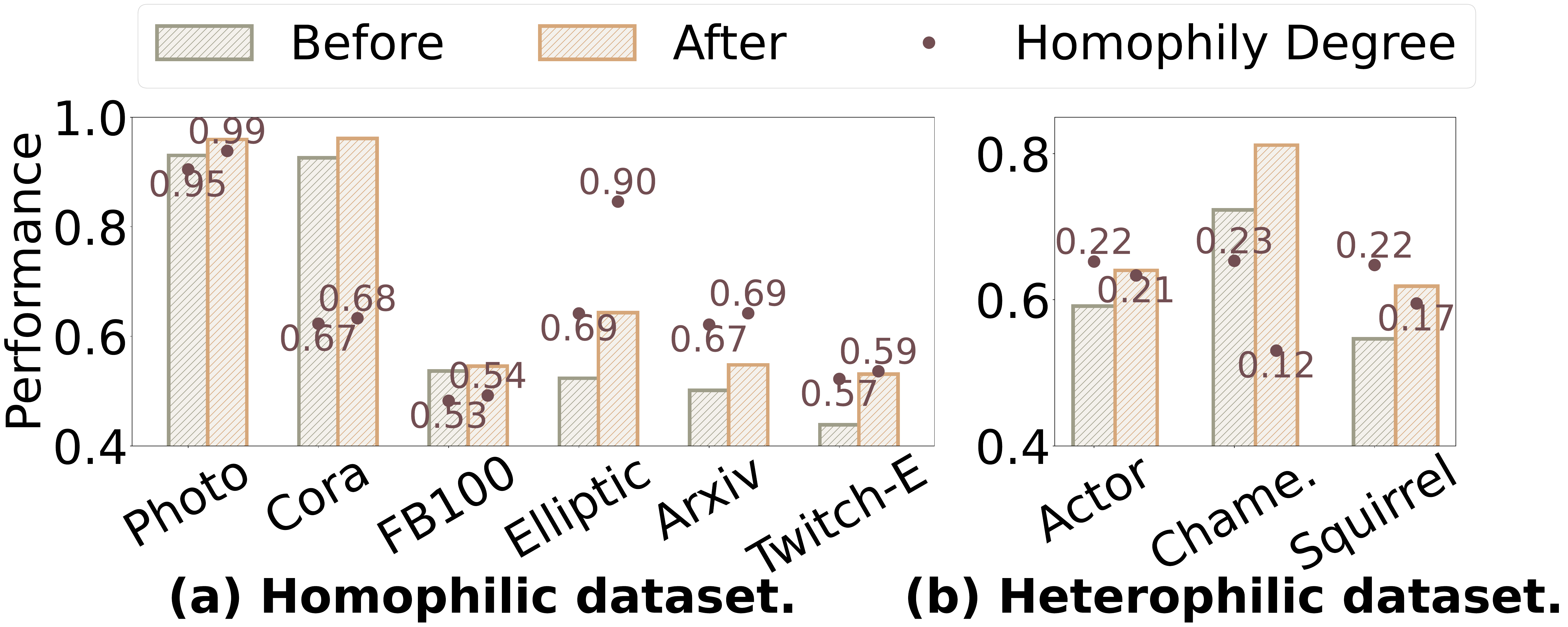}
    \end{minipage}
\caption{
Performance in bars and homophily degree in numbers before and after GrapHoST. As the homophily degree changes, the performance of GrapHoST is improved.
}
\label{fig:effect_drop}
\end{figure}

This experiment aims to evaluate the relationship between performance and the change in the homophily degree of test graphs after applying edge filtering in GrapHoST. As shown in Figure~\ref{fig:effect_drop}, in \textbf{homophilic graphs}, such as Photo, Elliptic, and Arxiv, \textbf{GrapHoST effectively enhances the average homophily degree} of the test graphs based on the homophily predictor's prediction and results in significant performance improvements compared to the original test graph. Similarly, for \textbf{heterophilic graphs, GrapHoST effectively reduces the average homophily degree}, leading to considerable performance gains.

\subsubsection{\textbf{Effectiveness of Homophily-weight Test Graph Construction}}
\begin{figure}[!h]
\centering
\includegraphics[width=1\linewidth]{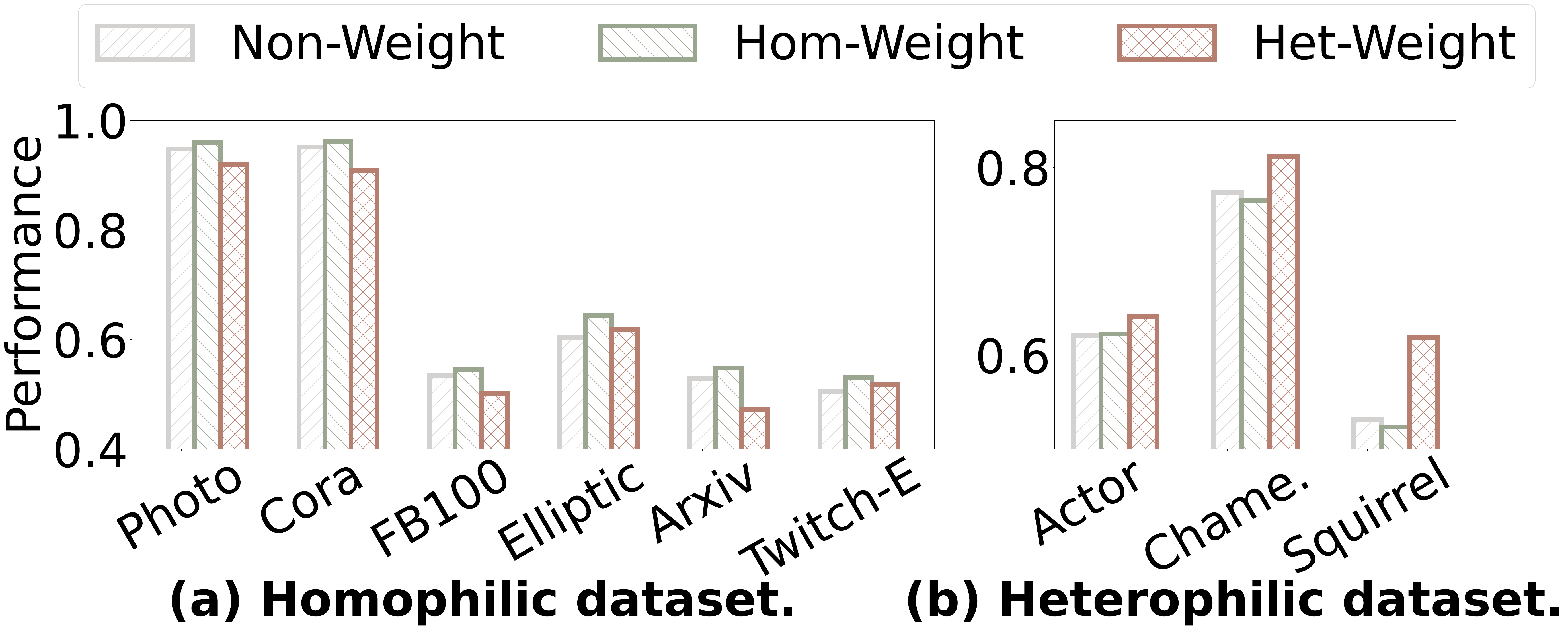} 
\caption{Performance under different test edge homophily weights. Using weight with (a) homophily scores in homophilic graphs or (b) heterophily scores in heterophilic graphs improves performance.}
\label{fig:reweight}
\end{figure}

This experiment evaluates different edge weights on test graphs: \textit{Non-Weight} without edge weight, \textit{Hom-Weight} with graphs weighted by homophily scores, and \textit{Het-Weight} with graphs weighted by heterophily scores. As shown in Figure~\ref{fig:reweight}, homophily-weighted graph construction improves GNN performance on homophilic graphs while using heterophily scores decreases performance. For heterophilic graphs, constructing the homophily-based test graphs via reweighting the edges by heterophily scores significantly improves performance. Therefore, the \textbf{homophily-based reweighting is effective for test graphs}.

\subsubsection{\textbf{Imbalanced Learning of Homophily Predictor}}
\label{sec:loss}
During training, the homophily predictor faces a significant imbalance between homophilic and heterophilic edges. To evaluate the impact, we compare predictors trained with standard BCE loss and weighted BCE (WBCE) loss. As shown in Table~\ref{tab:loss}, GrapHoST with WBCE performs better than the base BCE loss in the imbalanced homophily types of edges.

\begin{table}[!h]
    \centering
    \caption{Effectiveness of weighted binary cross-entropy loss.}
    \resizebox{\linewidth}{!}{
        \begin{tabular}{l|cccc}
            \toprule
            \textbf{Method} & \textbf{Photo} & \textbf{Cora} & \textbf{Elliptic} & \textbf{Arxiv} \\
            \midrule
            \toprule
            ERM (GCN) & 93.02$\pm$1.15 & 92.58$\pm$1.49 & 52.35$\pm$6.50 & 50.14$\pm$4.16 \\
            \midrule
            GrapHoST + BCE & 93.29$\pm$0.66 & 93.05$\pm$0.98 & 53.15$\pm$4.34 & 49.87$\pm$3.51 \\
            \rowcolor{gray!30}GrapHoST + WBCE & \textbf{95.94$\pm$0.85} & \textbf{96.12$\pm$0.47} & \textbf{64.32$\pm$3.80} & \textbf{54.78$\pm$2.83} \\
            \bottomrule
        \end{tabular}
    }
    \label{tab:loss}
\end{table}

\subsubsection{\textbf{Accuracy of Homophily Predictor on Test Graphs}}
In Table~\ref{tab:predictor}, we demonstrate that the homophily predictor exhibits strong homophily prediction performance on test graphs even with quality issues. The homophily predictor shows a \textbf{high prediction ROC-AUC score} and the \textbf{homophily degree of homophily graphs is increased while decreased for heterophily graphs}.

\begin{table}[!h]
    \centering
    \caption{Effectiveness of the homophily predictor. The ROC AUC score of the predictor is evaluated due to the imbalance of edge classes in graphs. The homophily degree ($HD.$) is also reported before and after applying GrapHoST based on the homophily predictor. All scores are in percentage (\%)}
    \resizebox{1\linewidth}{!}{
        \begin{tabular}{l|ccc|c}
            \toprule
            & \multicolumn{3}{c|}{Hom.}&Het.\\
            & \textbf{Photo} & \textbf{Cora} & \textbf{Elliptic} & \textbf{Chameleon}  \\
            \midrule
            \midrule
            \textbf{ROC-AUC} & 91.94±0.57 & 96.31±0.32 & 77.95±0.37 & 68.79±1.12  \\
            \midrule
            \textbf{$HD.$ Before} & 95.46 & 67.30 & 69.17 & 22.01 \\
            \textbf{$HD.$ After} & 99.14±0.96 & 68.29±0.27 & 90.06±0.17 & 17.98±1.14  \\
            \midrule
            \textbf{$\Delta HD.$} & +3.68 & +0.99 & +20.89 & -4.03 \\
            \bottomrule
        \end{tabular}
    }\label{tab:predictor}
    \vspace{-0.2cm}
\end{table}

\subsubsection{\textbf{Embedding Visualisation before and after Graph Transformation}}
\label{sec:umap}
\begin{figure}[!h]
\vspace{-0.4cm}
\centering
     \begin{minipage}[t]{1\linewidth}
        \centering
        \includegraphics[width=\linewidth]{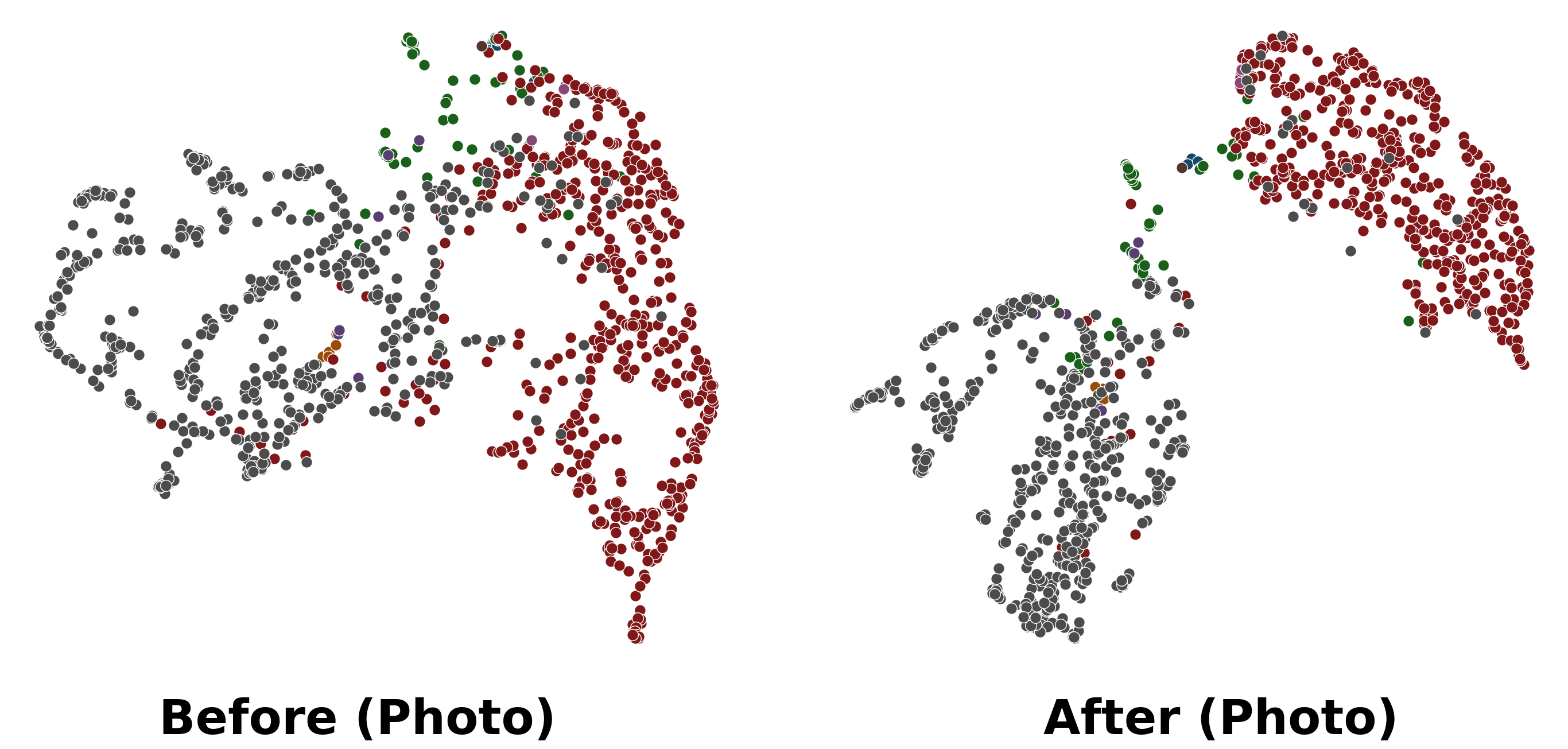}
    \end{minipage}

    \begin{minipage}[t]{\linewidth}
        \centering
        \includegraphics[width=\linewidth]{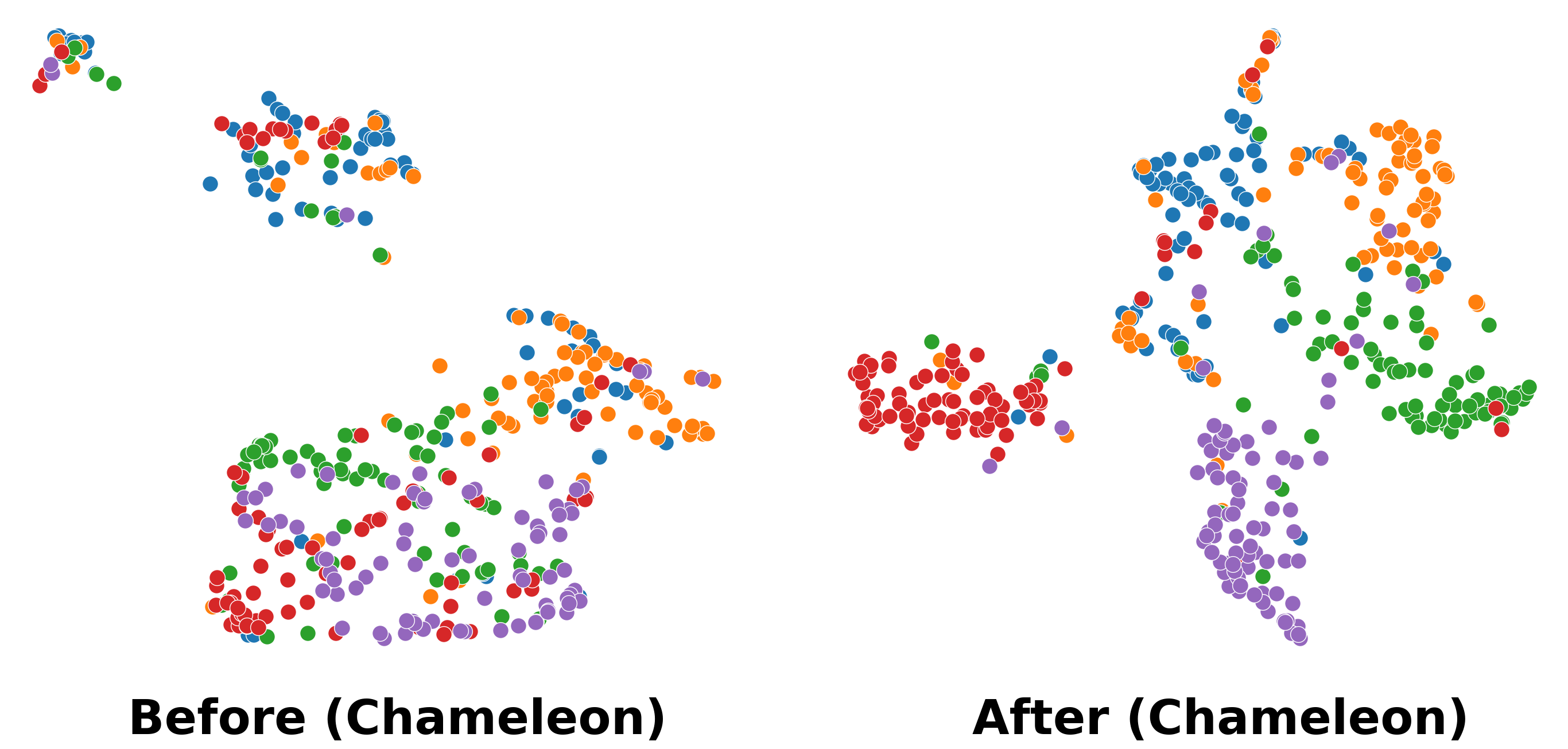}
    \end{minipage}
\caption{
UMAP visualisation of node embeddings on test graphs before and after GrapHoST. Colours represent classes. GrapHoST enhances class separation as in Theorem~\ref{thm:homo-imp}.}
\label{fig:umap}
\end{figure}

To validate the effectiveness of GrapHoST in improving test graph quality and GNN test-time performance, UMAP~\cite{UMAP} visualisations are performed to compare a fixed GNN classifier output embeddings on test graphs with data quality problems before and after applying GrapHoST. As shown in Figure~\ref{fig:umap}, the GNN encoded embeddings for test graphs after applying GrapHoST exhibit a clearer class separation, verifying that \textbf{GrapHoST effectively enhances the class separation of the GNN output embeddings.}

\subsection{Complexity and Efficiency Analysis}
This experiment analyses the complexity of the proposed GrapHoST method. During testing, the computational overhead arises from calculating homophily scores for each edge, as GrapHoST performs fine-grained edge-level transformations on the test graph. This results in a complexity of $O(E)$, where $E$ is the test edge number.

In addition, experiments on test-time graph transformation time and space efficiency are provided in Table~\ref{tab:time}, in comparison with existing test-time transformation methods. \textbf{GrapHoST demonstrates superior time and space efficiency, especially on large-scale graphs} such as Arxiv, due to its highly efficient graph structural transformation performed at a fine-grained edge level.
\begin{table}[!h]
    \centering
    \caption{Efficiency comparison. Each cell contains: test-time graph transformation time in seconds (s)/ GPU peak memory usage (GB)/ test-time performance (\%).}
    \resizebox{\linewidth}{!}{
        \begin{tabular}{l|cccc}
            \toprule
            \textbf{s/GB/\%} & \textbf{Photo} & \textbf{Cora} & \textbf{Elliptic} & \textbf{Arxiv} \\
            \midrule
            \midrule
            GP & 5.53/3.81/92.81 & 4.73/2.76/86.75 & 7.96/2.15/57.99 & 29.47/14.28/45.41 \\
            GTrans & 0.59/1.50/93.51 & 0.37/1.98/95.47 & 0.61/1.47/56.52 & 2.64/3.98/49.92 \\
            \midrule
            \rowcolor{gray!30}\textbf{Ours} & \textbf{0.06/1.20/95.94} & \textbf{0.30/0.06/96.12} & \textbf{0.31/0.19/64.32} & \textbf{0.07/2.56/54.78} \\
            \bottomrule
        \end{tabular}
    }
    \label{tab:time}
\end{table}

\subsection{Hyperparameter Sensitivity Analysis}
\label{sec:hyper}
\begin{figure}[!h]
\centering
\includegraphics[width=\linewidth]{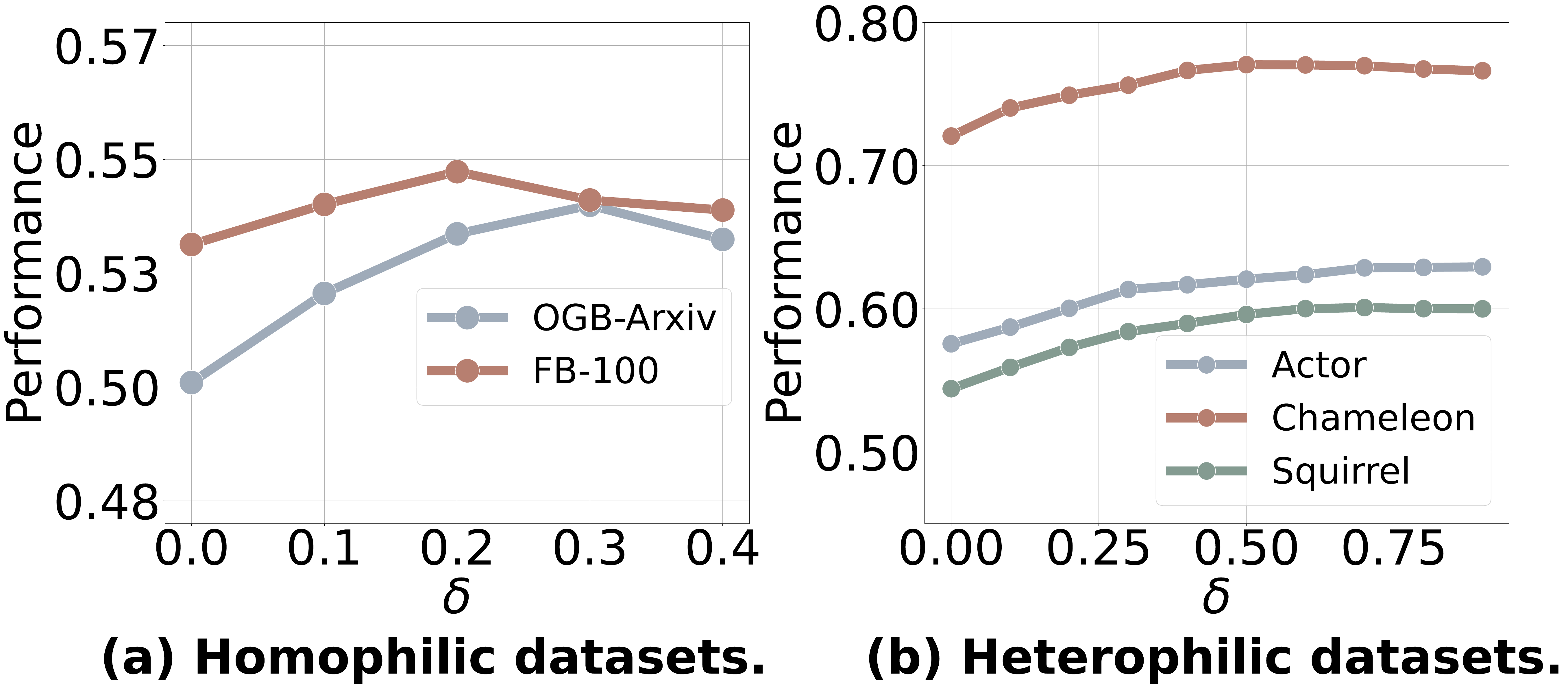} 
\caption{
Performance against different edge filtering ratios. (a) For homophilic datasets, as the number of dropped edges increases, the fixed GNN performance initially improves before subsequently declining. (b) For heterophilic datasets, as the number of dropped edges increases, the test-time performance of the fixed classifier continues to improve.
}
\label{fig:parameter1}
\end{figure}

This experiment examines the sensitivity of the key hyperparameter of GrapHoST, edge filtering ratio $\delta$ in Eq.~\eqref{eq:filtering}. By default, $\delta$ is set to 0.3 and it can yield satisfactory results. As shown in Figure~\ref{fig:parameter1}, for Arxiv and FB100, increasing $\delta$ initially improves GNN performance, but performance declines once $\delta$ grows further, suggesting that excessive edge filtering discards structural information. For heterophilic datasets, performance rises steadily with $\delta$ and plateaus at high filtering rates, as greater edge filtering preserves more pronounced homophily-based structural patterns. \textbf{This demonstrates the stability of the choice of different $\delta$ in GrapHoST.}

\section{Related Work}
\label{sec:relate}
\subsection{\textbf{Homophily and Heterophily on Graphs}}
Homophily and heterophily are fundamental properties of graph structure, and prior studies have shown that they can influence the training procedure of GNNs~\cite{Geom-GCN, zhu2020beyond, GOAL, bo2021beyond, GPR, yan2022two, luan2021heterophily, li2022finding, ACM, ma2021homophily, DCGC}. Unlike existing training-time homophily-based approaches, GrapHoST enhances GNN performance by leveraging \textbf{homophily-based properties in test graphs, without altering the classifier}.

\subsection{\textbf{Test-time Graph Data Quality Issues}}
Data quality issues in test graphs have presented a significant challenge to the real-world deployment of GNNs~\cite{GTRANS}. For example, GNNs often exhibit sub-optimal performance when there is a misalignment between the training and test graph distributions~\cite{EERM,zhu2021shift,liu2022confidence,chen2022invariance,buffelli2022sizeshiftreg,alsa,gold,tntood,puma,cat,caselink,casegnn}. Moreover, graph structures from different sources can be disrupted by human errors, outliers, and structure attacks at test-time, leading to degradation in the performance of the pre-trained GNNs~\cite{zugner2018adversarial,li2021adversarial}.

\subsection{\textbf{Test-time Training for Graphs}}
To address various test-time data quality issues, test-time training was proposed to update the model at test time. Recent approaches~\cite{GraphTTA, GT3, HomoTTT, LLMTTT, Matcha} have made notable progress by adapting or retraining the model during inference. For example, Matcha~\cite{Matcha} adjusts the hop aggregation parameter in GNNs during test-time to address the structural quality issues in the test graphs. \textbf{These methods are model-centric and primarily focus on enhancing model robustness instead of modifying the input test graphs.}

\subsection{\textbf{Test-time Graph Transformation}}
A few studies have been proposed to address test data quality issues by directly modifying the test graphs~\cite{GTRANS, GP}. GTrans~\cite{GTRANS} pioneers test-time graph transformation via contrastive learning on test graphs in a global view. GraphPatcher~\cite{GP} trains an auxiliary node generator model to patch neighbour nodes into test graphs. \textbf{However, critical homophily-based properties in test graph structures have been overlooked, limiting the potential for further test-time performance gains. To bridge the gap, GrapHoST systematically explores and exploits these properties, revealing their significant potential in enhancing GNN test-time performance.} Built on both empirical and theoretical analysis, GrapHoST introduces a fine-grained edge-level transformation on test graphs, marking a fundamental departure from prior methods.

\section{Conclusion}
In this paper, GrapHoST is proposed as a novel test-time graph structural transformation method that enhances robust node classification by modifying the test graph structure based on homophily-related properties in the test graphs. Both empirical and theoretical results indicate that homophily-based properties in test graphs, specifically increasing homophily in homophilic test graphs or decreasing it in heterophilic test graphs, enhance the performance of a fixed pre-trained GNN for robust node classification. Extensive experiments on nine benchmark datasets with various test data quality issues further validate both its effectiveness and efficiency.

\section{Acknowledgements}
This research has been supported by Australian Research Council Discovery Projects (DP230101196 and DE250100919).

\clearpage
\appendix
\bibliographystyle{ACM-Reference-Format}
\bibliography{sample-base}

\clearpage
\appendix
\appendix
\twocolumn[
\begin{center}
    \Huge \textbf{Supplementary Material for GrapHoST}
\end{center}
\vspace{1em}
]
\section{Supplementary Material Overview}
In the Appendix, additional supplementary material to the main paper is provided. The structure is as follows:

\begin{itemize}[left=0em, itemsep=0pt, topsep=0pt, parsep=0pt, partopsep=0pt]
\item The reproducibility statement is provided in Appendix~\ref{sec:reproduce}.
\item Detailed definition of embeddings under GNN operations is provided in Appendix~\ref{sec:embedding}.
\item Proof for Lemma 1 and Lemma 2 is provided in Appendix~\ref{sec:proof-lemma}.
\item Proof for Theorem~\ref{thm:homo-imp} and Theorem~\ref{thm:heter-imp} in the binary classification case is provided in Appendix~\ref{sec:proof-binary}, with: 
\begin{itemize}[left=0.2em, itemsep=0pt, topsep=0pt, parsep=0pt, partopsep=0pt]
 \item a more detailed proof for the relaxation of degree invariance in Appendix~\ref{sec:relax-degree}.
 \item a more detailed analysis for the relaxation of class balance in Appendix~\ref{sec:relax-boundary}
\end{itemize}
\item Proof for Theorem~\ref{thm:homo-imp} and Theorem~\ref{thm:heter-imp} generalised to multi-class classification case is provided in Appendix~\ref{sec:proof-multi}.
\item More experimental results and further analysis are provided in Appendix~\ref{app:further-exp}, the structure is:
    \begin{itemize}[left=0.2em, itemsep=0pt, topsep=0pt, parsep=0pt, partopsep=0pt]
        \item Appendix~\ref{sec:backbone} provides more results for GrapHoST using different backbone models for the homophily predictor.
        \item Appendix~\ref{sec:di} provides more results for GrapHoST and baseline methods on directed graph datasets.
        \item Appendix~\ref{sec:static} provides more results for GrapHoST adopted on static GNNs compared with random edge-dropping.
        \item Appendix~\ref{sec:quality} provides more details on data quality issues.
        \item Appendix~\ref{sec:clean-heter} provides more results for GrapHoST, improving base GNNs on clean heterophilic graphs by different splits.
        \item Appendix~\ref{sec:filter} provides more analysis on the filtered heterophilic datasets following the data preprocessing process in prior work~\cite{heter}.
        \item Appendix~\ref{sec:homo-shift} provides more analysis on the homophily pattern shift existed in the datasets with cross-domain or temporal evolution issues.
    \end{itemize}
\item More detailed related work is provided in Appendix~\ref{app:related}, the structure is:
    \begin{itemize}[left=0.2em, itemsep=0pt, topsep=0pt, parsep=0pt, partopsep=0pt]
        \item Appendix~\ref{app:quality} provides more detailed related work for graph data quality.
        \item Appendix~\ref{app:operation_1} provides more detailed related work for test time training methods.
        \item Appendix~\ref{app:operation_2} provides more detailed related work for test time graph transformation methods.
        \item Appendix~\ref{app:homophily} provides more detailed related work for homophily and heterophily.
    \end{itemize}
    
\item Further discussions on related fields are provided in Appendix~\ref{app:discuss}, the structure is:
    \begin{itemize}[left=0.2em, itemsep=0pt, topsep=0pt, parsep=0pt, partopsep=0pt]
        \item Appendix~\ref{app:gsl} discusses the relation of GrapHoST with training-time graph structure learning methods.
        \item Appendix~\ref{app:link} discusses the relation of GrapHoST with link prediction and graph clustering methods.
        \item Appendix~\ref{app:node_attribute} discusses the effectiveness of GrapHoST in addressing node attribute shift through homophily-based structural transformation.
        \item Appendix~\ref{app:test-time} analyses how GrapHoST relates to existing test-time methods with respect to the definition of test time.
    \end{itemize}
\end{itemize}

\section{Reproducibility Statement}
\label{sec:reproduce}

To promote reproducible research, we summarise our efforts as:
\begin{itemize}[left=0em, itemsep=0pt, topsep=0pt, parsep=0pt, partopsep=0pt]
\item \textbf{Open Source.} The code has been released at\\ \href{https://github.com/YanJiangJerry/GrapHoST}{\textcolor{purple}{https://github.com/YanJiangJerry/GrapHoST}}

\item \textbf{Baselines} We adopt the baseline methods from Tent~\cite{Tent}, EERM~\cite{EERM}, GTrans~\cite{GTRANS}, GraphPatcher~\cite{GP}, and Matcha~\cite{Matcha}. We carefully tune the hyperparameters according to the instructions in their papers to get the highest results for a fair comparison.

\item \textbf{Datasets} We utilise nine publicly available benchmark datasets. including:
\begin{itemize}[left=0.2em, itemsep=0pt, topsep=0pt, parsep=0pt, partopsep=0pt]
    \item Cora~\cite{yang2016revisiting} is a citation network in which nodes represent academic publications, and edges denote citation relationships.
    \item Amazon-Photo~\cite{shchur2018pitfalls} is a co-purchasing network where nodes correspond to products, and edges indicate frequent co-purchases.
    \item Twitch-Explicit~\cite{rozemberczki2021multi} (Twitch-E) consists of seven networks, each representing a community of Twitch users, where nodes correspond to individual users and edges capture mutual friendships.
    \item Facebook-100~\cite{lim2021new} (FB-100) comprises 100 snapshots of Facebook friendship networks from 2005, with each network representing a specific American university, where nodes correspond to Facebook users and edges represent friendships.
    \item Elliptic~\cite{Finance} is a temporal dataset consisting of 49 sequential graph snapshots, where each snapshot represents a network of Bitcoin transactions. Nodes correspond to individual transactions, and edges denote payment flows.
    \item OGB-Arxiv~\cite{OGB} contains 169,343 computer science papers from 40 subject areas, forming a citation network. The objective is to predict the subject area of a given paper.
    \item Actor~\cite{Geom-GCN} is an actor-only induced sub-graph extracted from the film-director-actor-writer network (Tang et al., 2009). Nodes represent actors, and an edge between two nodes indicates co-occurrence on the same Wikipedia page. Node attributes correspond to keywords extracted from the associated Wikipedia pages.
    \item Chameleon and Squirrel~\cite{Geom-GCN} are two page-page networks derived from Wikipedia. In these datasets, nodes represent web pages, edges signify mutual hyperlinks and node features encode informative nouns extracted from Wikipedia pages.
\end{itemize}

The process of constructing data quality issues follows the data preprocessing procedures from previous research that aims to study the impact of data quality issues~\cite{EERM, GTRANS}. More details on data quality issues are provided in the Appendix~\ref{sec:quality}.

\item \textbf{Methodology.} Our GrapHoST framework is fully documented in the method section. In addition, we provide a detailed pseudo-code in the Algorithm~\ref{alg:training} in the main paper. As a plug-and-play method, GrapHoST is built upon GTrans~\cite{GTRANS} with its feature transformation employed as a preprocess step.

\item \textbf{Model Training.} The homophily predictor is trained on a binary classification training set for homophily edges, where edges are labelled based on whether they connect nodes of the same class. The predictor is optimised by Adam~\cite{ADAM} optimiser using the proposed WBCE loss, as detailed in the main paper, with an in-depth analysis of the WBCE loss provided in Appendix~\ref{sec:loss}. Batch normalisation is adopted in line with prior study~\cite{GTRANS}, and early stopping is implemented to avoid over-fitting. 

\item \textbf{Backbones.} The default GNN backbone for the homophily predictor is GCN~\cite{gcn}, with its hidden state and number of layers set to match the fixed GNN classifier (e.g., a hidden dimension of 32 with 2 layers), ensuring a fair and consistent experiment without additional computational overhead. The implementation of the GNN backbone model is built upon the open-source work GTrans~\cite{GTRANS}. The results of adopting more backbone models, such as  GAT~\cite{gat}, GraphSAGE~\cite{sage}, GPR~\cite{GPR}, H2GCN~\cite{h2gcn} and LINKX~\cite{linkx}, are analysed in Appendix~\ref{sec:backbone}.

\item \textbf{Hyperparameter.} After training the homophily predictor, the only hyperparameter that needs to be searched is the confidence threshold, denoted as $\delta$, which is determined by evaluating the GNN performance on the validation graphs. The search space is discussed in the main paper. Once the homophily predictor is trained and the optimal edge filtering ratio is determined, the graph structural transformation can be performed at test time.

\item \textbf{Evaluation Metrics.} The GNN test-time performance is evaluated using accuracy and F1 macro scores, and the performance of the trained homophily predictor is evaluated by ROC AUC scores as in Section~\ref{sec:effectiveness}.

\item \textbf{Device.} NVIDIA V100 GPU with 32GB memory is used for all the experiments.
\end{itemize}

\section{Detailed Proof for Binary Classification Case}
\label{sec:binary}

\subsection{Linear Separability under GNNs}
\label{sec:embedding}
To thoroughly assess the effectiveness of GNNs, linear classifiers with the largest margin based on $ \{\mathbf{h}_i, i \in \mathcal{V}\} $ are analysed without incorporating non-linearity, following previous work~\cite{insights, linear, ma2021homophily}. Specifically, the GNN message aggregation process $ \mathbf{h}_i = \frac{1}{\text{deg}(i)} \sum_{j \in \mathcal{N}(i)} \mathbf{x_j} $ is considered, where $ \text{deg}(i) $ denotes the degree of the node $ i $. For a graph generated by CSBM, denoted as:
\begin{equation}
\begin{aligned}
\label{eq:csbm}
\mathcal{G} \sim \text{CSBM}(\boldsymbol{\mu_1}, \boldsymbol{\mu_2}, p, q).
\end{aligned}
\end{equation}

The labels of the node neighbours are independently sampled from neighbourhood distributions, which can be defined by:
\begin{equation}
\begin{aligned}
\mathcal{D} = \begin{cases} 
\left[\frac{p}{p+q}, \frac{q}{p+q}\right] & \text{for } c_1, \\ 
\left[\frac{q}{p+q}, \frac{p}{p+q}\right] & \text{for } c_2.
\end{cases}
\end{aligned}
\end{equation}

Based on these neighbourhood distributions, the GNN output embeddings follow Gaussian distributions:
\begin{equation}
\begin{aligned}
\label{eq:h1}
\mathbf{h}_i \sim \begin{cases} 
\mathcal{N} \left(\frac{p\boldsymbol{\mu_1} + q\boldsymbol{\mu_2}}{p+q}, \frac{\mathbf{I}}{\text{deg}(i)} \right) & \text{for } c_1, \\ 
\mathcal{N} \left(\frac{q\boldsymbol{\mu_1} + p\boldsymbol{\mu_2}}{p+q}, \frac{\mathbf{I}}{\text{deg}(i)} \right) & \text{for } c_2. 
\end{cases}
\end{aligned}
\end{equation}

When the homophily degree of the graph is changed, the initialised node features according to definition~\ref{eq:csbm} are kept unchanged. Specifically, when $p$ is changed to $p'$ and $q$ is changed to $q'$, the GNN output embeddings are as follows:
\begin{equation}
\begin{aligned}
\label{eq:h2}
\mathbf{h}'_i \sim \begin{cases} 
\mathcal{N} \left(\frac{p'\boldsymbol{\mu_1} + q'\boldsymbol{\mu_2}}{p'+q'}, \frac{\mathbf{I}}{\text{deg}(i)} \right) & \text{for } c_1, \\ 
\mathcal{N} \left(\frac{q'\boldsymbol{\mu_1} + p'\boldsymbol{\mu_2}}{p'+q'}, \frac{\mathbf{I}}{\text{deg}(i)} \right) & \text{for } c_2. 
\end{cases}
\end{aligned}
\end{equation}

\setcounter{lemma}{0}
\subsection{Detailed Proof for Lemmas}
\label{sec:proof-lemma}

\begin{tcolorbox}[colframe=gray!60, colback=gray!10, coltitle=black, boxsep=0pt, left=2pt, right=2pt, top=2pt, bottom=2pt]

\begin{lemma}
The vectors $ (\mathbb{E}_{c_1}[\mathbf{h}'_i], \mathbb{E}_{c_2}[\mathbf{h}'_i]) $ and $ (\mathbb{E}_{c_1}[\mathbf{h}_i], \mathbb{E}_{c_2}[\mathbf{h}_i]) $ share the same midpoint: $ \mathbf{m} = \frac{\boldsymbol{\mu_1} + \boldsymbol{\mu_2}}{2} $. 
\end{lemma}
\end{tcolorbox}

\begin{proof}
For the original graph:
\begin{equation}
\begin{aligned}
\mathbb{E}_{c_1}[\mathbf{h}_i] &= \frac{p\boldsymbol{\mu_1} + q\boldsymbol{\mu_2}}{p+q}, \\
\mathbb{E}_{c_2}[\mathbf{h}_i] &= \frac{q\boldsymbol{\mu_1} + p\boldsymbol{\mu_2}}{p+q}.
\end{aligned}
\end{equation}

For the transformed graph:
\begin{equation}
\begin{aligned}
\mathbb{E}_{c_1}[\mathbf{h}'_i] &= \frac{p'\boldsymbol{\mu_1} + q'\boldsymbol{\mu_2}}{p'+q'}, \\
\mathbb{E}_{c_2}[\mathbf{h}'_i] &= \frac{q'\boldsymbol{\mu_1} + p'\boldsymbol{\mu_2}}{p'+q'}.
\end{aligned}
\end{equation}

The expectation midpoints in both graphs are:
\begin{equation}
\begin{aligned}
\mathbf{m} &= \frac{\mathbb{E}_{c_1}[\mathbf{h}_i] + \mathbb{E}_{c_2}[\mathbf{h}_i]}{2} = \frac{\boldsymbol{\mu_1} + \boldsymbol{\mu_2}}{2}, \\
\mathbf{m}' &= \frac{\mathbb{E}_{c_1}[\mathbf{h}'_i] + \mathbb{E}_{c_2}[\mathbf{h}'_i]}{2} = \frac{\boldsymbol{\mu_1} + \boldsymbol{\mu_2}}{2}.
\end{aligned}
\end{equation}
It is clear that the midpoints are the same.
\end{proof}

\begin{tcolorbox}[colframe=gray!60, colback=gray!10, coltitle=black, boxsep=0pt, left=2pt, right=2pt, top=2pt, bottom=2pt]
\begin{lemma}
The differences $ \mathbb{E}_{c_1}[\mathbf{h}'_i] - \mathbb{E}_{c_2}[\mathbf{h}'_i] $ and $ \mathbb{E}_{c_1}[\mathbf{h}_i] - \mathbb{E}_{c_2}[\mathbf{h}_i] $ align in the same direction: $ \mathbf{o} = \frac{\boldsymbol{\mu_1} - \boldsymbol{\mu_2}}{\|\boldsymbol{\mu_1} - \boldsymbol{\mu_2}\|_2} $. 
\end{lemma}
\end{tcolorbox}

\begin{proof}
For the original graph:
\begin{equation}
\begin{aligned}
\mathbb{E}_{c_1}[\mathbf{h}_i] - \mathbb{E}_{c_2}[\mathbf{h}_i] &= \frac{(p - q)(\boldsymbol{\mu_1} - \boldsymbol{\mu_2})}{p+q}.
\end{aligned}
\end{equation}

For the transformed graph:
\begin{equation}
\begin{aligned}
\mathbb{E}_{c_1}[\mathbf{h}'_i] - \mathbb{E}_{c_2}[\mathbf{h}'_i] &= \frac{(p' - q')(\boldsymbol{\mu_1} - \boldsymbol{\mu_2})}{p'+q'},
\end{aligned}
\end{equation}
where $p, q, p', q'$ are all scalars. Therefore, both graphs' expectation differences share the same direction:
\begin{equation}
\begin{aligned}
\mathbf{o} &= \frac{\boldsymbol{\mu_1} - \boldsymbol{\mu_2}}{\|\boldsymbol{\mu_1} - \boldsymbol{\mu_2}\|_2}.
\end{aligned}
\end{equation}
\end{proof}

\setcounter{theorem}{0}
\subsection{Detailed Proof of Theorem~\ref{thm:homo-imp}}
\label{sec:proof-binary}

\begin{tcolorbox}[
  colframe={rgb,1:red,0.75;green,0.60;blue,0.45},  
  colback={rgb,1:red,0.98;green,0.96;blue,0.93},   
  coltitle=black,
  boxsep=0pt,
  left=2pt,
  right=2pt,
  top=2pt,
  bottom=2pt
]
\begin{theorem}
    (\textbf{Node Classification on Homophilic Test Graph under a Fixed Classifier}). For node $i$ in a homophilic test graph $\mathcal{G}^{Hom}_{Te} \sim \text{CSBM}(\boldsymbol{\mu_1}, \boldsymbol{\mu_2}, p, q), \text{ where } (p > q)$, with a fixed classifier defined by the decision boundary $\mathcal{B}$ as in Proposition~\ref{prop:db}, if $\mathcal{G}^{Hom}_{Te}$ is transformed into $\mathcal{G'}^{Hom}_{Te} \sim \text{CSBM}(\boldsymbol{\mu_1}, \boldsymbol{\mu_2}, p', q')$ with a higher homophily degree $(p' > p) \text{ and } (q' < q)$, the misclassification probability of $\mathcal{B}$ on $\mathcal{G}'^{Hom}_{Te}$ is lower than $\mathcal{G}^{Hom}_{Te}$. 
\end{theorem}
\end{tcolorbox}

\begin{proof}
For a node $i \in c_1$ in a binary classification task, the probabilities of misclassification can be defined as:
\begin{equation}
\begin{aligned}
\mathcal{P}_{\text{mis}, c_1} = \mathbb{P}(\mathbf{\mathbf{W}}^\top \mathbf{h}_i + \mathbf{b} \leq 0) \quad \text{for } i \in c_1,
\end{aligned}
\end{equation}
where $\mathbf{W}$ and $\mathbf{b}$ are the weight and bias of the classifier model.

The L2 distance between the expected value of the GNN output embeddings to the decision boundary is:
\begin{equation}
\begin{aligned}
\label{eq:dist}
d &= \frac{1}{2} \|\mathbb{E}_{c_1}[\mathbf{h}] - \mathbb{E}_{c_2}[\mathbf{h}]\|_2 \\
&= \frac{1}{2} \left\| \frac{(p - q)(\boldsymbol{\mu_1} - \boldsymbol{\mu_2})}{p + q} \right\|_2 \\
&= \frac{1}{2} \cdot\frac{|p - q|}{p + q} \cdot \|\boldsymbol{\mu_1} - \boldsymbol{\mu_2}\|_2.
\end{aligned}
\end{equation}

For homophilic graphs where nodes tend to connect the same class neighbours, $p > q$ and $|p - q| = p - q$. Thus,
\begin{equation}
\begin{aligned}
d_{hom} &= \frac{1}{2} \cdot \frac{p - q}{p + q} \cdot \|\boldsymbol{\mu_1} - \boldsymbol{\mu_2}\|_2 \\
&= \frac{a}{2} \cdot \frac{p - q}{p + q}, 
\end{aligned}
\end{equation}
where $ \|\boldsymbol{\mu_1} - \boldsymbol{\mu_2}\|_2$ can be denoted as a constant $a$. During the homophily-based graph structural transformation, as the node features are unchanged, $\boldsymbol{\mu_1}$ and $\boldsymbol{\mu_2}$ are fixed, thus the distance between the expected value of the GNN output embeddings of the homophily-based transformed graph $\mathcal{G'}^{Hom}_{Te}$ is:
\begin{equation}
\begin{aligned}
d'_{hom} = \frac{a}{2} \cdot \frac{p' - q'}{p' + q'},
\end{aligned}
\end{equation}
where $p' > p > 0$ and $q > q' > 0$ because the homophily degree has increased from $\mathcal{G}^{Hom}_{Te}$ to $\mathcal{G'}^{Hom}_{Te}$. 

To evaluate the misclassification probability, the distance between the embeddings from $\mathcal{G}^{Hom}_{Te}$ in Eq.~\eqref{eq:h1} and those from $\mathcal{G'}^{Hom}_{Te}$ in Eq.~\eqref{eq:h2} can be directly compared under the assumption that the variance $\frac{\mathbf{I}}{\text{deg}(i)}$ remains unchanged, indicating the node degree of $\mathcal{G}^{Hom}_{Te}$ and $\mathcal{G'}^{Hom}_{Te}$ remain invariance. \textbf{Note that when such an assumption does not hold, additional constraints can be imposed to ensure the validity of the theorems, which are rigorously proved in Appendix~\ref{sec:relax-degree}.} 

Therefore, comparing the distances:
\begin{equation}
\begin{aligned}
d'_{hom} - d_{hom} &= \frac{a}{2} \left( \frac{p' - q'}{p' + q'} - \frac{p - q}{p + q} \right) \\
&= \frac{a \left( (p' - q')(p + q) - (p - q)(p' + q') \right)}{2(p' + q')(p + q)} \\
&= \frac{a \left( p'q - q'p \right)}{(p' + q')(p + q)} \\
&> 0.
\end{aligned}
\end{equation}

The larger the distance from the expected embeddings to the decision boundary, the lower the probability of misclassification:
\begin{equation}
\begin{aligned}
\mathcal{P}_{\text{mis}, c'_1} < \mathcal{P}_{\text{mis}, c_1}.
\end{aligned}
\end{equation}

Therefore, for homophilic graphs, the fixed classifier at test-time defined by the decision boundary $\mathcal{B}$ has a lower probability of misclassifying the GNN output embeddings $\mathbf{h}'_i$ on the test graph $\mathcal{G'}^{Hom}_{Te}$ after increasing the homophily degree than the GNN output embeddings $\mathbf{h}_i$ on the original test graph $\mathcal{G}^{Hom}_{Te}$. 
\end{proof}

\setcounter{theorem}{1}
\subsection{Detailed Proof of Theorem~\ref{thm:heter-imp}}
\label{sec:proof-binary-heter}

\begin{tcolorbox}[
  colframe={rgb,1:red,0.75;green,0.60;blue,0.45},  
  colback={rgb,1:red,0.98;green,0.96;blue,0.93},   
  coltitle=black,
  boxsep=0pt,
  left=2pt,
  right=2pt,
  top=2pt,
  bottom=2pt
]
\begin{theorem}
(\textbf{Node Classification on Heterophilic Test Graph under a Fixed Classifier}). For node $i$ in a heterophilic test graph $\mathcal{G}^{Het}_{Te} \sim \text{CSBM}(\boldsymbol{\mu_1}, \boldsymbol{\mu_2}, p, q), \text{ where } (p < q)$, with a fixed classifier defined by the decision boundary $\mathcal{B}$ as in Proposition~\ref{prop:db}, if $\mathcal{G}^{Het}_{Te}$ is transformed into $\mathcal{G'}^{Het}_{Te} \sim \text{CSBM}(\boldsymbol{\mu_1}, \boldsymbol{\mu_2}, p', q')$ with a lower homophily degree $(p' < p) \text{ and } (q' > q)$, the misclassification probability of $\mathcal{B}$ on $\mathcal{G}'^{Het}_{Te}$ is lower than $\mathcal{G}^{Het}_{Te}$. 
\end{theorem}
\end{tcolorbox}
\vspace{-0.1cm}

\begin{proof}
From Eq.~\eqref{eq:dist}, for heterophilic graphs where nodes tend to connect the different class nodes, $p < q$ and $|p - q| = q - p$. Thus,
\begin{equation}
\begin{aligned}
d_{het} &= \frac{1}{2} \cdot \frac{q - p}{p + q} \cdot \|\boldsymbol{\mu_1} - \boldsymbol{\mu_2}\|_2. \\
&= \frac{a}{2} \cdot \frac{q - p}{p + q}.
\end{aligned}
\end{equation}

During the homophily-based graph structural transformation, as the node features are unchanged, $\boldsymbol{\mu_1}$ and $\boldsymbol{\mu_2}$ are fixed, thus the distance between the expected value of the GNN output embeddings of the homophily-based transformed graph $\mathcal{G'}^{Het}_{Te}$ is:
\begin{equation}
\begin{aligned}
d'_{het} = \frac{a}{2} \cdot \frac{q' - p’}{p' + q'},
\end{aligned}
\end{equation}
where $q' > q > 0$ and $p > p' > 0$ because the homophily degree has decreased. Comparing two distances:
\begin{equation}
\begin{aligned}
d'_{het} - d_{het} &= \frac{a}{2} \left( \frac{q' - p’}{p' + q'} - \frac{q - p}{p + q} \right) \\
&= \frac{a \left( (q' - p’)(p + q) - (q - p)(p' + q') \right)}{2(p' + q')(p + q)} \\
&= \frac{a \left( q'p - p'q \right)}{(p' + q')(p + q)} \\
&> 0.
\end{aligned}
\end{equation}

The larger the distance from the expected embeddings to the decision boundary, the lower the probability of misclassification:
\begin{equation}
\begin{aligned}
\mathcal{P}_{\text{mis}, c'_1} < \mathcal{P}_{\text{mis}, c_1}.
\end{aligned}
\end{equation}

Therefore, for heterophilic graphs, the fixed classifier at test-time defined by the decision boundary $\mathcal{B}$ has a lower probability of misclassifying the GNN output embeddings $\mathbf{h}'_i$ on the test graph $\mathcal{G'}^{Het}_{Te}$ after decreasing the homophily degree than the GNN output embeddings $\mathbf{h}_i$ on the original test graph $\mathcal{G}^{Het}_{Te}$, which completes the proof.
\end{proof}

\subsection{Relaxation of Degree Invariance Assumption}
\label{sec:relax-degree}
\begin{proof}
In this section, the degree invariance assumption in Appendix~\ref{sec:proof-binary} is relaxed, and we aim to prove that additional constraints can be imposed to ensure the validity of Theorem~\ref{thm:homo-imp} and Theorem~\ref{thm:heter-imp} by incorporating the degree into the analysis of the misclassification probability.

Given the distributions of node embeddings from Appendix~\ref{sec:embedding}:
\begin{equation}
\begin{aligned}
\mathbf{h}_i \sim \begin{cases} 
\mathcal{N} \left(\frac{p\boldsymbol{\mu_1} + q\boldsymbol{\mu_2}}{p+q}, \frac{\mathbf{I}}{\text{deg}(i)} \right) & \text{for } c_1, \\ 
\mathcal{N} \left(\frac{q\boldsymbol{\mu_1} + p\boldsymbol{\mu_2}}{p+q}, \frac{\mathbf{I}}{\text{deg}(i)} \right) & \text{for } c_2. 
\end{cases}
\end{aligned}
\end{equation}
According to Lemma 1, the optimal decision boundary is fixed at the midpoint under the assumption that the class is balanced:
\begin{equation}
\label{eq:boundary}
\mathcal{B} = \frac{\frac{p \boldsymbol{\mu_1} + q \boldsymbol{\mu_2}}{p + q} + \frac{q \boldsymbol{\mu_1} + p \boldsymbol{\mu_2}}{p + q}}{2}
= \frac{\boldsymbol{\mu_1} + \boldsymbol{\mu_2}}{2},
\end{equation}
\textbf{Note that the case where the class balance assumption is relaxed is also analysed in Appendix~\ref{sec:relax-boundary}.}

In the following, we will rigorously prove the misclassification probability for class $ c_1 $ since a similar step can be applied for class $ c_2 $. For class $ c_1 $, the correct classification occurs when $ x \leq \mathcal{B} $. The decision boundary $\mathcal{B}$ can be projected onto a standard normal distribution space by standardising with respect to the distribution mean and variance for the embedding of class $ c_1 $:
\begin{equation}
\begin{aligned}
\mathcal{P}_{\text{correct}, c_1} &= \mathcal{P}(x \leq \mathcal{B} \mid x \sim \mathcal{N} \left(\frac{p \boldsymbol{\mu_1} + q \boldsymbol{\mu_2}}{p + q}, \sigma_1 \right))\\ &= \Phi\left( \frac{\|\mathcal{B} - \frac{p \boldsymbol{\mu_1} + q \boldsymbol{\mu_2}}{p + q}\|}{\sigma_1} \right).
\end{aligned}
\end{equation}
Substituting $ \mathcal{B} = \frac{\boldsymbol{\mu_1} + \boldsymbol{\mu_2}}{2} $, we get:
\begin{equation}
\begin{aligned}
\mathcal{P}_{\text{correct}, c_1} = \Phi\left( \frac{|q - p| \cdot \|\boldsymbol{\mu_1} - \boldsymbol{\mu_2}\|}{2(p + q)\sigma_1} \right).
\end{aligned}
\end{equation}

\begin{equation}
\begin{aligned}
\mathcal{P}_{\text{mis}, c_1} &= 1 - \mathcal{P}_{\text{correct}, c_1}\\
&= 1- \Phi\left( \frac{|q - p| \cdot \|\boldsymbol{\mu_1} - \boldsymbol{\mu_2}\|}{2(p + q)\sigma_1} \right)\\
&= \Phi\left( -\frac{|p - q| \cdot \|\boldsymbol{\mu_1} - \boldsymbol{\mu_2}\|}{2(p + q)\sigma_1} \right).
\end{aligned}
\end{equation}

Assume the number of nodes in class $ c_1 $ are $n_1$ and $n_2$, therefore, the average degree for node in class $ c_1 $ is $p(n_1 - 1) + qn_2$. Therefore, the standard variance for embedding in class $ c_1 $ is
\begin{equation}
\sigma_1 = \frac{\mathbf{I}}{\text{deg}(i)} = \frac{1}{\sqrt{p(n_1 - 1) + qn_2}}.
\end{equation}

By approximating $n_1 - 1 \approx n_1$, we get the average degree as $p n_1 + q n_2$. The standard deviation for embedding in class $c_1$ is:

\begin{equation}
\sigma_1 = \frac{1}{\sqrt{p n_1 + q n_2}}.
\end{equation}

Set \( a = \|\boldsymbol{\mu_1} - \boldsymbol{\mu_2}\| \) is a constant representing the Euclidean distance between \(\boldsymbol{\mu_1}\) and \(\boldsymbol{\mu_2}\).
The misclassification probabilities are:

\begin{equation}
\mathcal{P}_{\text{mis}, c_1} = \Phi\left( -\frac{a |p - q| \sqrt{p n_1 + q n_2}}{2(p + q)} \right),
\end{equation}

Similarly, for $G'$:
\begin{equation}
\mathcal{P}_{\text{mis}, c'_1} = \Phi\left( -\frac{a|p' - q'| \sqrt{p' n_1 + q' n_2}}{2(p' + q')} \right),
\end{equation}

\subsubsection{Homophilic Graph Cases}

When the graph is a homophilic graph, with the probability of homophilic edge connectivity being greater than that of heterophilic connectivity ($p > q$ and $p' > q'$):

\begin{equation}
\begin{aligned}
\mathcal{P}_{\text{mis}, c_1} &= \Phi\left( -\frac{a(p - q) \sqrt{p n_1 + q n_2}}{2(p + q)} \right).
\end{aligned}
\end{equation}

Similarly, for $G'$:

\begin{equation}
\begin{aligned}
\mathcal{P}_{\text{mis}, c'_1} &= \Phi\left( -\frac{a(p' - q') \sqrt{p' n_1 + q' n_2}}{2(p' + q')} \right).
\end{aligned}
\end{equation}

Comparing the misclassification probability:

\begin{equation}
\begin{aligned}
&\mathcal{P}_{\text{mis}, c_1} - \mathcal{P}_{\text{mis}, c'_1} \\
&= \Phi\left( -\frac{a(p - q)\sqrt{p n_1 + q n_2}}{2(p + q)} \right) - \Phi\left( -\frac{a(p' - q') \sqrt{p' n_1 + q' n_2}}{2(p' + q')} \right).
\end{aligned}
\end{equation}

Based on the monotonicity of \(\Phi\), we have:
\begin{equation}
\begin{aligned}
&\mathcal{P}_{\text{mis}, c_1} - \mathcal{P}_{\text{mis}, c'_1} \\
&\equiv a \cdot \left( \frac{(p' - q')\sqrt{p' n_1 + q' n_2}}{2(p' + q')} - \frac{(p - q)\sqrt{p n_1 + q n_2}}{2(p + q)} \right) \\
&= a \cdot \frac{(p' - q')(p + q)\sqrt{p' n_1 + q' n_2} - (p - q)(p' + q')\sqrt{p n_1 + q n_2}}{2(p + q)(p' + q')}.
\end{aligned}
\end{equation}

As $a > 0$ and $2(p + q)(p' + q') > 0$, the sign of $P - P'$ depends on the numerator:

\begin{equation}
(p' - q')(p + q)\sqrt{p' n_1 + q' n_2} > (p - q)(p' + q')\sqrt{p n_1 + q n_2}.
\end{equation}

Under such a constraint:
\begin{equation}
\mathcal{P}_{\text{mis}, c_1} > \mathcal{P}_{\text{mis}, c'_1},
\end{equation}

which implies that Theorem~\ref{thm:homo-imp} holds under the given constraint when the degree invariance assumption is relaxed.

\subsubsection{Heterophilic Graph Cases}
When the graph is a heterophilic graph, with the probability of homophilic edge connectivity being less than that of heterophilic connectivity ($p < q$ and $p' < q'$):

\begin{equation}
\begin{aligned}
\mathcal{P}_{\text{mis}, c_1} &= \Phi\left( -\frac{(q - p)a\sqrt{p n_1 + q n_2}}{2(p + q)} \right).
\end{aligned}
\end{equation}

For \( G' \):

\begin{equation}
\begin{aligned}
\mathcal{P}_{\text{mis}, c'_1} &= \Phi\left( -\frac{(q' - p')a\sqrt{p' n_1 + q' n_2}}{2(p' + q')} \right).
\end{aligned}
\end{equation}

Comparing the misclassification probability:

\begin{equation}
\begin{aligned}
&\mathcal{P}_{\text{mis}, c_1} - \mathcal{P}_{\text{mis}, c'_1} \\
&= \Phi\left( -\frac{(q - p)a\sqrt{p n_1 + q n_2}}{2(p + q)} \right) - \Phi\left( -\frac{(q' - p')a\sqrt{p' n_1 + q' n_2}}{2(p' + q')} \right).
\end{aligned}
\end{equation}

Based on the monotonicity of \(\Phi\), we have:
\begin{equation}
\begin{aligned}
&\mathcal{P}_{\text{mis}, c_1} - \mathcal{P}_{\text{mis}, c'_1} \\
&\equiv a \cdot \left( \frac{(q' - p')\sqrt{p' n_1 + q' n_2}}{2(p' + q')} - \frac{(q - p)\sqrt{p n_1 + q n_2}}{2(p + q)} \right) \\
&= a \cdot \frac{(q' - p')(p + q)\sqrt{p' n_1 + q' n_2} - (q - p)(p' + q')\sqrt{p n_1 + q n_2}}{2(p + q)(p' + q')}.
\end{aligned}
\end{equation}

As \( a > 0 \) and \( 2(p + q)(p' + q') > 0 \), the sign of \( P - P' \) depends on the numerator:

\begin{equation}
(q' - p')(p + q)\sqrt{p' n_1 + q' n_2} > (q - p)(p' + q')\sqrt{p n_1 + q n_2}.
\end{equation}

Under such a constraint:
\begin{equation}
\mathcal{P}_{\text{mis}, c_1} > \mathcal{P}_{\text{mis}, c'_1},
\end{equation}

which implies that Theorem~\ref{thm:heter-imp} holds under the given constraint when the degree invariance assumption is relaxed. Therefore, Theorem~\ref{thm:homo-imp} and Theorem~\ref{thm:heter-imp} are valid by introducing the above constraints upon relaxing degree invariance, which completes the proof.
\end{proof}

\subsection{Relaxation of Balanced Class Assumption}
\label{sec:relax-boundary}

The optimal decision boundary in Lemma 1 establishes that the optimal decision boundary lies fixed at the midpoint between the expectations of two classes under the assumption of balanced classes. In this section, the position of the optimal decision boundary is analysed upon the relaxation of the class balance assumption.

When the number of nodes in each class is imbalanced, the optimal decision boundary can be derived based on the number of nodes in each class, $ n_1 $ and $ n_2 $. Set $ g_1(x)$ and $ g_2(x) $ to be Gaussian likelihoods for the correct classification probability for $c_1$ and $c_2$. $ \pi(c_1) $ and $ \pi(c_2) $ denote the class prior based on the number of nodes:
\begin{equation}
\pi(c_1) = \frac{n_1}{n_1 + n_2}, \quad \pi(c_2) = \frac{n_2}{n_1 + n_2}.
\end{equation}

The optimal decision boundary is determined by:
\begin{equation}
\pi(c_1) g_1(x) = \pi(c_2) g_2(x),
\end{equation}
where \( g_1(x) \) and \( g_2(x) \) are the Gaussian likelihood functions for classes \( c_1 \) and \( c_2 \), respectively:
\begin{equation}
g_1(x) = \frac{1}{\sqrt{2 \pi \sigma_1^2}} e^{-\frac{(x - \mu_1)^2}{2 \sigma_1^2}}, \quad 
g_2(x) = \frac{1}{\sqrt{2 \pi \sigma_2^2}} e^{-\frac{(x - \mu_2)^2}{2 \sigma_2^2}}.
\end{equation}

By taking the log-likelihood ratio:
\begin{equation}
\ln \frac{\pi(c_1)}{\pi(c_2)} + \frac{(x - \mu_2)^2}{2 \sigma_2^2} - \frac{(x - \mu_1)^2}{2 \sigma_1^2} = 0.
\end{equation}

Therefore, the final decision boundary \( \mathcal{B} \) is:
\begin{equation}
\mathcal{B} = \frac{\sigma_1^2 \mu_2 - \sigma_2^2 \mu_1}{\sigma_1^2 - \sigma_2^2} + \frac{\ln (n_2 / n_1)}{\sigma_1^2 - \sigma_2^2}.
\end{equation}

When the variance of the class $c_1$ and $c_2$ is the same (\( \sigma_1 = \sigma_2 = \sigma \)), the decision boundary can be simplified to:
\begin{equation}
\mathcal{B} = \frac{\mu_1 + \mu_2}{2} + \frac{\ln (n_2 / n_1)}{2 \sigma^2}.
\end{equation}
This new form of decision boundary introduces an additional term related to the class instance number, compared to the decision boundary under the class balance assumption in Eq.~\ref{eq:boundary}.

\section{Proof for Multi-class Classification Cases}
\label{sec:proof-multi}
Theorem~\ref{thm:homo-imp} and Theorem~\ref{thm:heter-imp} in multi-class classification can be generalised from the proof for binary classification in Appendix~\ref{sec:proof-binary}.

\begin{proof}
In a $s$-class CSBM, the nodes in the generated graphs are partitioned into $s$ distinct sets, denoted as $c_1, \dots, c_s$. Similarly, the edges between nodes are generated based on two probabilities: an intra-class connection probability $p$ and an inter-class connection probability $q$. For any pair of nodes, if both nodes belong to the same class, an edge is established with probability $p$; if the nodes belong to different classes, the edge probability is $q$. Each node $i$ on the graph is assigned an initial feature vector ${\bf x}_i \in \mathbb{R}^l$, which is sampled from a Gaussian distribution: ${\bf x}_i \sim N(\boldsymbol{\mu}, {\bf I})$ with $\boldsymbol{\mu}$ representing the mean vector associated with the class of nodes $i$. In particular, $\boldsymbol{\mu}_v \in \mathbb{R}^l$ for nodes $i$ belonging to the class $c_v$. Moreover, the mean vectors for distinct classes are different, that is, $\boldsymbol{\mu}_v \neq \boldsymbol{\mu}_g$ for any pair of distinct classes $v, g \in \{1, \dots, s\}$. The Euclidean distance between the mean vectors of two classes can be denoted by a constant $d$, that is, $\| \boldsymbol{\mu}_v - \boldsymbol{\mu}_g \|_2 = d$ for all pairs $v, g \in \{1, \dots, s\}$.

For class $c_v$, the distribution of neighbourhood labels $\mathcal{D}_{c_v}$ can be represented by a vector in which the $v$-th component equals $\frac{p}{p + (s - 1)q}$, while all other components are equal to $\frac{q}{p + (s - 1)q}$. The analysed GNN operation is the same as binary classification cases:
\begin{equation}
\begin{aligned}
{\bf \mathbf{h}}_i = \frac{1}{\text{deg}(i)} \sum_{j \in \mathcal{N}(i)} {\bf x}_j,
\end{aligned}
\end{equation}
where $\text{deg}(i)$ denotes the degree of node $i$, and $\mathcal{N}(i)$ represents the set of neighbours of node $i$. For a node $i$ with label $c_v$, the embeddings after such an aggregation process follow a Gaussian distribution, which can be described by:
\begin{equation}
\begin{aligned}
    {\bf \mathbf{h}}_i \sim N\left(\frac{p \boldsymbol{\mu}_v + \sum_{j \neq v} q \boldsymbol{\mu}_j}{p + (s - 1)q}, \frac{{\bf I}}{\text{deg}(i)} \right), \text{ for } y_i = c_v. 
\end{aligned}
\end{equation}

The expected values for class $c_v$ after GNN encoding are:
\begin{equation}
\begin{aligned}
\mathbb{E}_{c_v}[{\bf \mathbf{h}}] = \frac{p \boldsymbol{\mu}_v + \sum_{j \neq v} q \boldsymbol{\mu}_j}{p + (s - 1)q}.
\end{aligned}
\end{equation}

For any two classes $ c_v $ and $ c_g $, the Euclidean distance between their expected values after the GNN operation can be calculated as:
\begin{equation}
\begin{aligned}
\label{eq:dist-multi}
    \|\mathbb{E}_{c_v}[{\bf \mathbf{h}}] - \mathbb{E}_{c_g}[{\bf \mathbf{h}}]\|_2 &= \left\| \frac{p \boldsymbol{\mu}_v + \sum_{j \neq v} q \boldsymbol{\mu}_j}{p + (s - 1)q} - \frac{p \boldsymbol{\mu}_g + \sum_{j \neq \mathbf{g}} q \boldsymbol{\mu}_j}{p + (s - 1)q} \right\|_2  \\
    &= \left\| \frac{p \boldsymbol{\mu}_v - p \boldsymbol{\mu}_g + q \sum_{j \neq v} \boldsymbol{\mu}_j - q \sum_{j \neq \mathbf{g}} \boldsymbol{\mu}_j}{p + (s - 1)q} \right\|_2  \\
    &= \left\| \frac{p (\boldsymbol{\mu}_v - \boldsymbol{\mu}_g) + q (\boldsymbol{\mu}_g - \boldsymbol{\mu}_v)}{p + (s - 1)q} \right\|_2  \\
    &= \frac{|p - q|}{p + (s - 1)q} \left\| \boldsymbol{\mu}_v - \boldsymbol{\mu}_g \right\|_2, 
\end{aligned}
\end{equation}
where class number $s$ and distance $ \|\boldsymbol{\mu_v} - \boldsymbol{\mu_g}\|_2 $ are constants. The rest of the proof for multi-class classification cases is the same as for binary classification from Eq.~\eqref{eq:dist}. Therefore, Theorem~\ref{thm:homo-imp} and Theorem~\ref{thm:heter-imp} are also valid in the multi-class classification case, which completes the proof.
\end{proof}

\section{Further Experimental Results}
\label{app:further-exp}

\begin{table*}[!t]
    \centering
    \caption{More results of GrapHost using homophily predictor (HOM) implemented by different backbones.}
    \resizebox{\linewidth}{!}{
        \begin{tabular}{c|l|cccccc|ccc}
            \toprule
            &\textbf{Method} & \textbf{Amz-Photo} & \textbf{Cora} & \textbf{FB-100} & \textbf{Elliptic} & \textbf{OGB-Arxiv} & \textbf{Twitch-E} & \textbf{Actor} & \textbf{Chameleon} & \textbf{Squirrel} \\
            \midrule
            \midrule
            &ERM (GCN) & 93.02$\pm$1.15 & 92.58$\pm$1.49 & 54.16$\pm$0.92 & 52.35$\pm$6.50 & 50.14$\pm$4.16 & 43.85$\pm$3.48 & 59.14$\pm$0.76 & 72.31$\pm$0.77 & 54.66$\pm$0.95 \\
            \midrule
            &HOM-MLP & 94.35$\pm$1.35 & 97.07$\pm$0.39 & 53.87$\pm$0.52 & 62.45$\pm$4.93 & 53.23$\pm$3.37 & 53.45$\pm$2.79 & 63.49$\pm$0.67 & 76.03$\pm$1.11 & 57.93$\pm$0.67 \\
            \midrule
            \multirow{4}{*}{\rotatebox{90}{Hom.}} &HOM-GCN & \color{teal}\underline{95.94$\pm$0.85} & 96.12$\pm$0.47 & \color{teal}\underline{54.68$\pm$0.76} & \color{purple}\textbf{64.32$\pm$3.80} & \color{purple}\textbf{54.78$\pm$2.83} & 53.52$\pm$1.49 & \color{teal}\underline{64.07$\pm$0.55} & \color{teal}\underline{81.17$\pm$0.48} & \color{teal}\underline{61.86$\pm$0.49} \\
            &HOM-GAT & 95.26$\pm$0.69 & 96.14$\pm$0.40 & \color{purple}\textbf{54.57$\pm$0.91} & 62.24$\pm$4.31 & \color{teal}\underline{53.29$\pm$3.26} & \color{purple}\textbf{53.99$\pm$0.56} & 63.39$\pm$0.68 & 79.45$\pm$0.75 & 59.80$\pm$0.77 \\
            &HOM-SAGE & \color{purple}\textbf{96.35$\pm$0.73} & \color{purple}\textbf{97.28$\pm$0.34} & 54.40$\pm$1.17 & \color{teal}\underline{63.61$\pm$4.32} & 52.62$\pm$3.41 & \color{teal}\underline{53.90$\pm$0.43} & 62.62$\pm$0.54 & 80.69$\pm$0.85 & 57.97$\pm$0.58 \\
            &HOM-GPR & 93.83$\pm$0.84 & 96.83$\pm$0.36 & 54.27$\pm$0.67 & 62.55$\pm$4.28 & 51.28$\pm$3.72 & 52.79$\pm$1.34 & 62.33$\pm$0.54 & 75.51$\pm$0.55 & 59.34$\pm$0.36 \\
            \midrule
            \multirow{2}{*}{\rotatebox{90}{Het.}} &HOM-H2GCN & 93.36$\pm$1.08 & 95.65$\pm$0.56 & 54.47$\pm$0.33 & 62.04$\pm$4.31 & 51.38$\pm$4.32 & 48.36$\pm$1.10 & 63.92$\pm$0.66 & \color{purple}\textbf{81.36$\pm$0.90} & \color{purple}\textbf{61.91$\pm$0.58} \\
            &HOM-LINKX & 94.53$\pm$0.96 & \color{teal}\underline{97.08$\pm$0.37} & 54.40$\pm$0.68 & 62.66$\pm$4.32 & 53.21$\pm$3.36 & 51.72$\pm$2.07 & \color{purple}\textbf{64.28$\pm$0.65} & 81.07$\pm$0.68 & 60.49$\pm$1.12 \\
            \midrule
        \end{tabular}
    }
    \label{tab:backbone_all}
\end{table*}

\begin{table}[!t]
    \centering
    \caption{Overall performance for directed graphs. Note that GTrans~\cite{GTRANS} adds reverse edges of existing edges thus gaining advantage by converting directed edges into undirected~\cite{OGB, GP}. GTrans* removes this restriction for fair comparisons.
    }
    \resizebox{\linewidth}{!}{
        \begin{tabular}{c|ccc|c}
            \toprule
            \textbf{Method} & \textbf{Elliptic} & \textbf{OGB-Arxiv} & \textbf{Twitch-E} & \textbf{Rank} \\
            \midrule\midrule
            ERM (GCN) & 49.81$\pm$1.29 & 39.15$\pm$1.49 & 39.69$\pm$3.66 & 6.33\\
            DropEdge & 52.05$\pm$1.94 & 40.28$\pm$1.01 & 48.00$\pm$2.49 & 3.67\\
            EERM & 51.30$\pm$3.35 & 34.05$\pm$0.96 & \color{teal}\underline{51.13$\pm$3.03} & 5.33\\
            Tent & 46.81$\pm$0.02 & 38.87$\pm$2.76 & 39.36$\pm$3.69 & 8.00\\
            FTTT & 47.67$\pm$0.77 & 27.97$\pm$3.09 & 38.35$\pm$3.21 & 9.00\\
            GraphPatcher & 46.97$\pm$0.22 & 37.93$\pm$1.77  & 27.47$\pm$2.95 & 8.67\\
            \midrule
            GTrans$^*$ & 54.40$\pm$1.92 & 32.79$\pm$0.97 & 40.14$\pm$3.07 & 5.67\\
            GTrans & \color{teal}\underline{54.80$\pm$2.02} & \color{purple}\underline{\textbf{40.71$\pm$1.03}} & 40.02$\pm$3.29 & \color{teal}\underline{3.00}\\
            \midrule
            \rowcolor{gray!30}
            \textbf{GrapHoST}  & \color{purple}\textbf{55.05$\pm$1.99} & 40.29$\pm$1.06 & \color{purple}\textbf{51.38$\pm$2.17} & \color{purple}\textbf{1.33}\\
            \bottomrule
        \end{tabular}
    }
    \label{tab:di}
\end{table}

\begin{table}[!t]
    \centering
    \caption{Applying GrapHoST for different static pre-trained GNNs on more datasets.}
    \resizebox{\linewidth}{!}{
        \begin{tabular}{l|ccccc}
            \toprule
            \textbf{Method} & \textbf{FB-100} & \textbf{Twitch-E} & \textbf{Actor} & \textbf{Chameleon} & \textbf{Squirrel} \\
            \midrule
            \midrule
            ERM (GCN) & 54.16$\pm$0.92 & 43.85$\pm$3.48 & 59.14$\pm$0.76 & 72.31$\pm$0.77 & 54.66$\pm$0.95 \\
            +Random & 53.75$\pm$1.16 & 44.76$\pm$4.42 & 60.17$\pm$1.53 & 68.62$\pm$0.98 & 52.84$\pm$1.17 \\
            \rowcolor{gray!30}\textbf{+GrapHoST} & 54.68$\pm$0.76 & 53.52$\pm$1.49 & 64.07$\pm$0.55 & 81.17$\pm$0.48 & 61.86$\pm$0.49 \\

            \midrule
            DropEdge & 53.67$\pm$0.88 & 48.93$\pm$1.70 & 60.16$\pm$0.81 & 73.36$\pm$0.63 & 56.97$\pm$1.05 \\
            +Random & 53.17$\pm$1.09 & 46.39$\pm$2.25 & 60.19$\pm$1.03 & 69.72$\pm$1.27 & 54.59$\pm$1.70 \\
            \rowcolor{gray!30}\textbf{+GrapHoST} & 54.37$\pm$0.75 & 50.08$\pm$2.64 & 65.40$\pm$0.63 & 82.31$\pm$0.60 & 63.84$\pm$1.25 \\

            \midrule
            EERM & 54.14$\pm$0.34 & 45.44$\pm$6.47 & 60.46$\pm$0.29 & 73.15$\pm$0.72 & 61.82$\pm$0.31 \\
            +Random & 53.98$\pm$0.77 & 45.52$\pm$5.27 & 59.24$\pm$0.41 & 69.09$\pm$0.80 & 59.07$\pm$0.46 \\
            \rowcolor{gray!30}\textbf{+GrapHoST} & 54.50$\pm$0.45 & 58.22$\pm$0.91 & 61.53$\pm$0.52 & 82.37$\pm$0.47 & 64.48$\pm$0.70 \\
            
            \bottomrule
        \end{tabular}
    }\label{tab:plug_2}
\end{table}

\subsection{GrapHoST upon Different Backbones}
\label{sec:backbone}
To assess the impact of various backbone model designs for the homophily predictor, GrapHoST is examined with different backbone models as the homophily predictor, including MLP, GNN models designed for homophilic graphs and GNN models tailored for heterophilic graphs, to analyse the effect of backbone model selection of the homophily predictor on the performance of GrapHoST on a fixed, well-trained GCN classifier. To conduct a fair comparison, the hyperparameters for each homophily predictor are tuned according to the validation graphs following the setup as in the main paper.

As presented in Table~\ref{tab:backbone_all}, adopting general GNN structures as homophily predictors, such as GraphSAGE, GAT, GCN and H2GCN, can enhance the test-time performance of the fixed GNN classifier. The effectiveness of different backbones varies across datasets due to their different representational capabilities on graphs. Generally, adopting a heterophilic GNN backbone as the homophily predictor achieves relatively higher results on heterophilic datasets compared to homophilic GNN backbones. To be noticed, \textbf{employing GCN as the backbone model for the homophily predictor in GrapHoST consistently improves test-time performance across all datasets.} In contrast, employing MLP as the homophily predictor leads to decreased performance on the FB-100 datasets. This is because MLP lacks the ability to capture graph structural information, making it ineffective at distinguishing edges based on homophily-related properties and unable to effectively adjust the homophily degree of the test graph as other GNN backbones. Therefore, the valuable connectivity patterns indicated by the homophily-based properties cannot be well preserved, leading to degraded GNN test-time performance.

\subsection{Overall Performance on Directed Graphs}
\label{sec:di}
As baseline methods~\cite{EERM, GTRANS} evaluate performance on directed Elliptic, OGB-Arxiv, and Twitch-E graphs, their results are also provided in Table~\ref{tab:di} and \textbf{GrapHoST still outperforms all baselines in directed graph settings.}

To be noticed, GTrans first transforms the directed test graph into an undirected edge space and then selects edges to add or remove based on this undirected representation. This process results in a transformed graph with added edges that correspond to the reverse edge of existing edges. Specifically, a direct conversion of the directed test graph to the undirected form increased GNN performance on OGB-Arxiv from 39.15±1.49 to 45.15±1.17. Therefore, the performance gain of GTrans on directed OGB-Arxiv can primarily be attributed to the conversion of the directed graph to an undirected one, rather than to the method's inherent contribution.

Given that simply converting a directed graph to an undirected one enhances performance, a fair comparison should be to test all methods on the converted undirected graph, or to refrain from adding reverse edges of the existing edges (converting to an undirected graph)~\cite{OGB, GP}. In the first scenario, GTrans* (32.79±0.97) is compared with the GrapHoST method (40.29±1.06), while in the second scenario, GTrans (40.71±1.03) is compared with the GrapHoST method that selects edges in the same undirected edge space, yielding an improved performance of 46.12±1.46. \textbf{Therefore, in both cases, the GrapHoST method still outperforms the state-of-the-art GTrans method on the directed OGB-Arxiv dataset.} Moreover, it is evident that the overall performance of the GNN classifiers trained on the directed graphs is significantly inferior to that trained on the undirected graphs in the overall table.

\subsection{Further Analysis on Static GNNs Compared with Random Edge Dropping}
\label{sec:static}
This section provides more results on applying GrapHoST to static GNNs trained by different training-time methods. Compared to randomly dropping edges, GrapHoST significantly improves the test-time performance of the static GNNs on all the provided datasets, demonstrating that GrapHoST is an effective plug-and-play method that can be easily integrated with existing methods to enhance their performance during test time. In contrast, randomly dropping edges with the same edge-dropping ratio leads to performance degradation in most datasets. For example, applying edge-dropping to ERM-trained GNN reduces the performance from 54.16 to 53.75, while employing GrapHoST increases the performance to 54.55, highlighting \textbf{ the effectiveness of the homophily-based graph structural transformation for static GNNs.}

\subsection{Further Details on Data Quality Issues}
\label{sec:quality}
To generate graphs with inherent data quality issues, the node attribute shift in Cora, Amazon-Photo and the newly introduced heterophilic datasets is introduced by keeping the training and test graphs the same, following the data quality preprocessing process in EERM~\cite{EERM} and GTrans~\cite{GTRANS} while adding synthetic spurious node attributes to the graph. In this case, the performance of the base GNN is solely affected by the node attribute shift between the training and test graphs. For heterophilic datasets, more detailed results on the clean graph by the original split~\cite{Geom-GCN} are discussed in Appendix~\ref{sec:clean-heter}. 

For cross-domain and temporal evolution scenarios in Twitch-E, FB-100, OGB-Arxiv and Elliptic, the data quality issues arise from selectively splitting the graph according to the domain or timestamp of the nodes following the split from previous work~\cite{EERM, GTRANS}, leading to significant changes in the degree of homophily between training and test graphs. A more detailed analysis of the homophily shift is provided in Appendix~\ref{sec:homo-shift}.

\subsection{Performance on Clean Heterophilic Graphs}
\label{sec:clean-heter}
To further evaluate the performance of GrapHoST without being influenced by inherent data quality issues in node attributes, further experiments are conducted on clean heterophilic graphs in an inductive setting following the split ratio provided in the prior study~\cite{Geom-GCN}. In this setup, the training and test graphs are not restricted to be the same as in EERM~\cite{EERM} and GTrans~\cite{GTRANS}, and no synthetic noisy feature is introduced. Therefore, data quality issues arise predominantly from the distribution shift between training and test graphs. Five GNN backbone classifiers, GCN~\cite{gcn}, GAT~\cite{gat}, GraphSAGE~\cite{sage}, GPR~\cite{GPR} and H2GCN~\cite{h2gcn}, are pre-trained and fixed during test time, with the homophily predictor in GrapHoST adopting the same backbone model as the classifiers.

As shown in Table~\ref{tab:heter}, the performance of the base GNN classifiers deteriorates due to changes in the data split and preprocessing procedures, in contrast to the node attribute shift configuration in Appendix~\ref{sec:quality}. As outlined in the problem definition in the main paper, GrapHoST is designed to improve the test-time performance of the base GNN classifiers, and the results across all datasets clearly demonstrate that \textbf{GrapHoST consistently enhances the test-time performance of baseline GNNs and outperforms other methods in various data quality scenarios, highlighting its robustness and effectiveness.}

\begin{figure*}[!t]
\centering
\includegraphics[width=\linewidth]{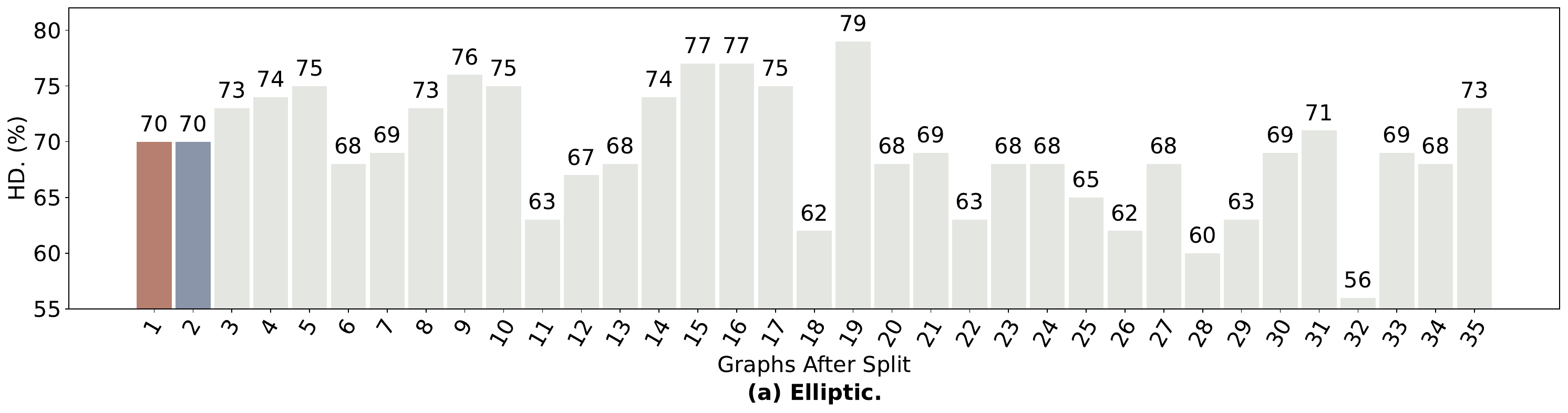} 

\begin{minipage}[t]{0.32\linewidth}
    \centering
    \includegraphics[width=\linewidth]{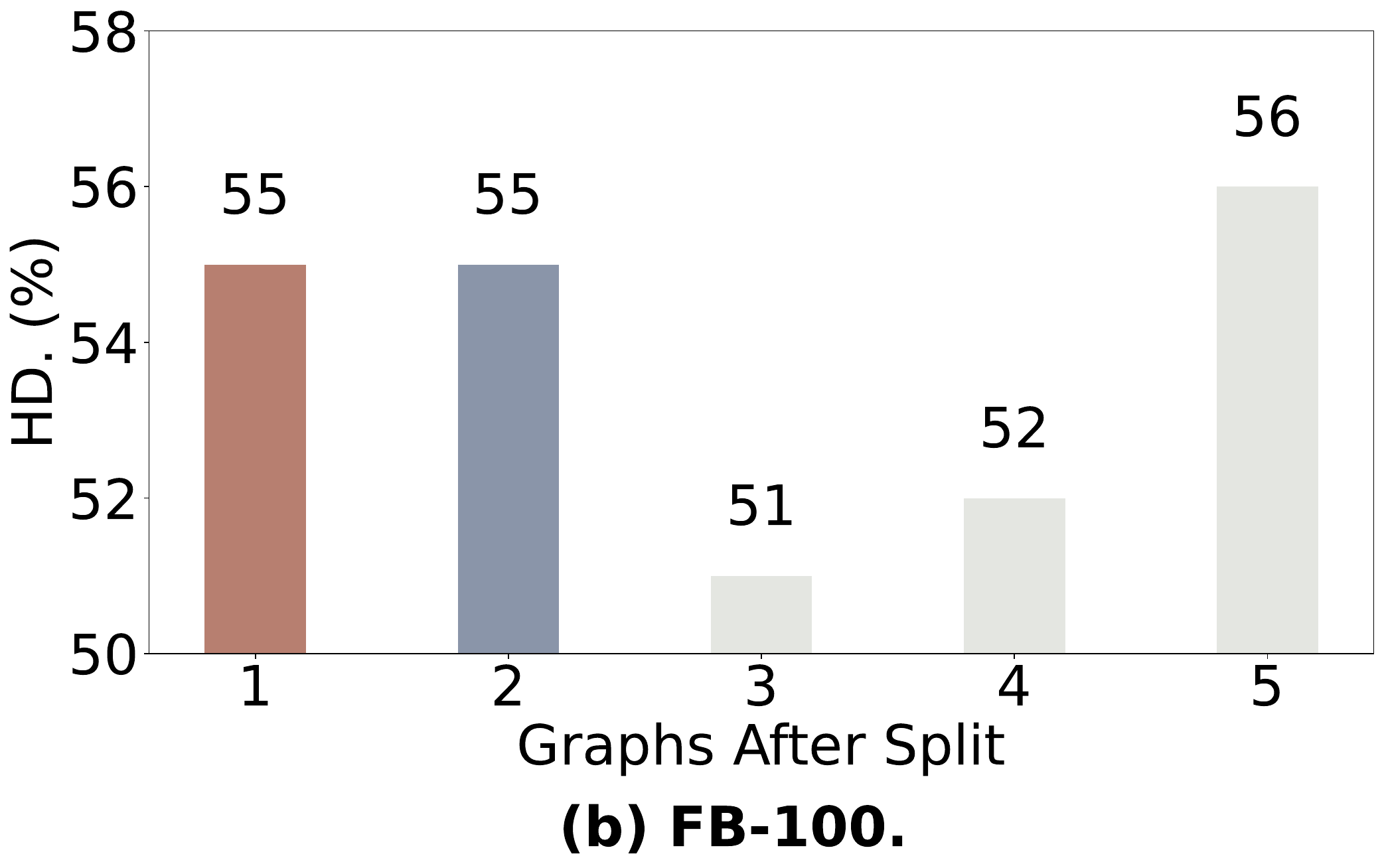}
    \vspace{-0.8cm}
\end{minipage}
\hfill
\begin{minipage}[t]{0.315\linewidth}
    \centering
    \includegraphics[width=\linewidth]{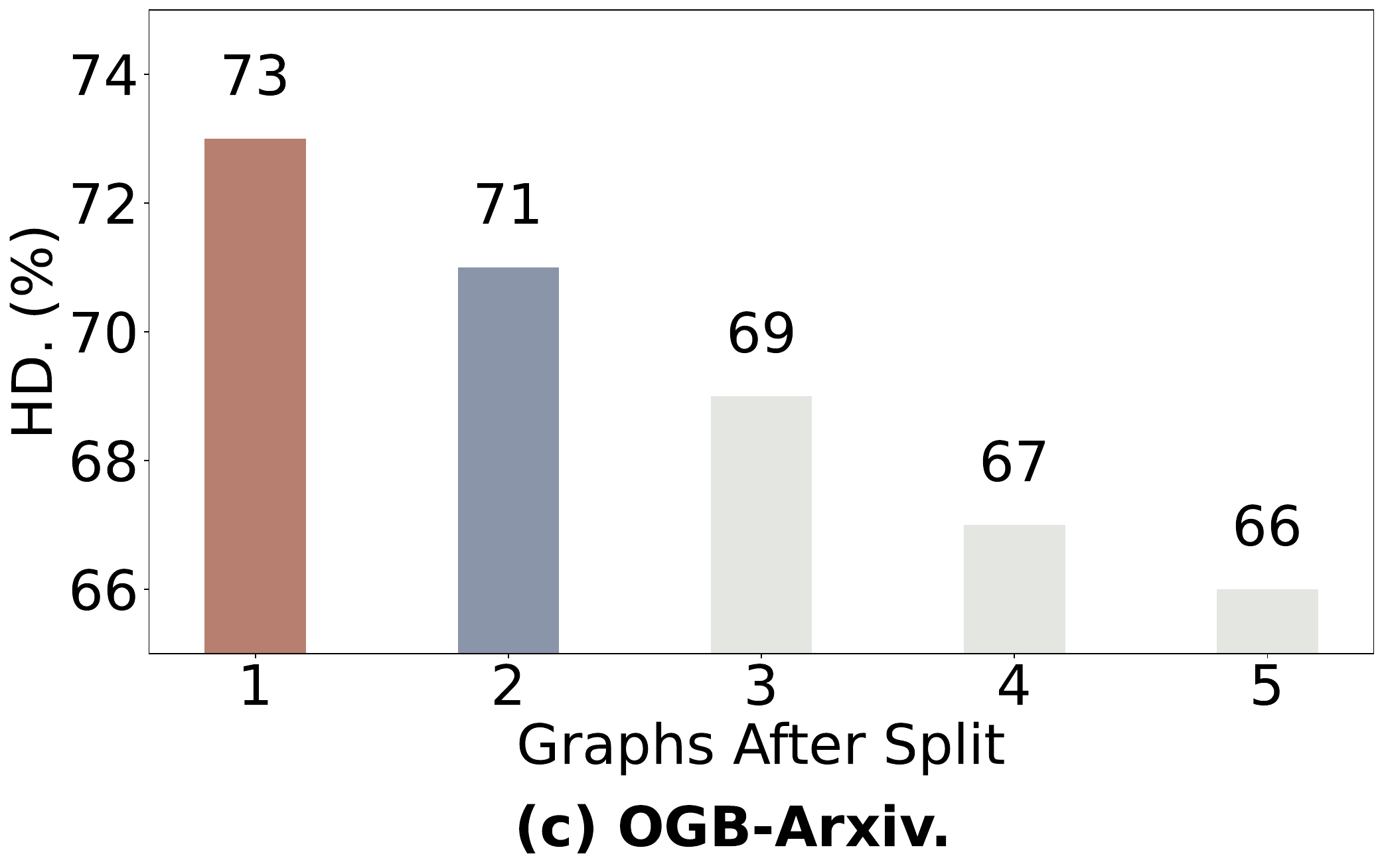}
    \vspace{-0.8cm}
\end{minipage}
\hfill
\begin{minipage}[t]{0.32\linewidth}
    \centering
    \includegraphics[width=\linewidth]{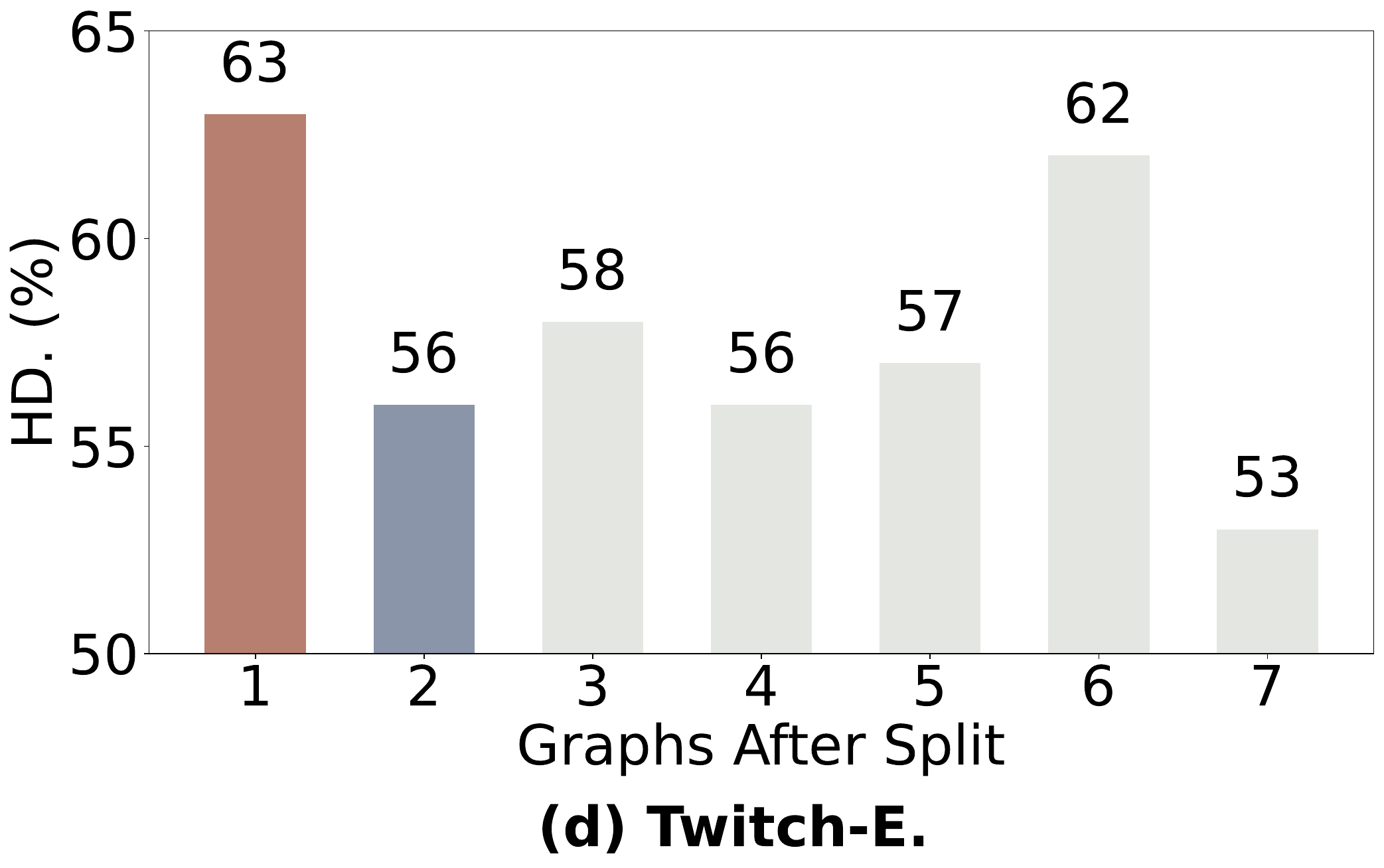}
    \vspace{-0.8cm}
\end{minipage}
\caption{Homophily shift across graphs after the split. The red and blue bars represent the average edge homophily degree for the training and validation graphs, while the other bars represent the average edge homophily degree for the test graphs.}
\vspace{-0.3cm}
\label{fig:homo_shift}
\end{figure*}

\begin{table}[!t]
    \centering
    \caption{Performance of GNNs trained on a clean heterophilic dataset by a different split in an inductive setting.}
  
    \resizebox{0.85\linewidth}{!}{
        \begin{tabular}{c|c|ccc}
            \toprule
            & \textbf{Method} & \textbf{Actor} & \textbf{Chameleon} & \textbf{Squirrel} \\
            \midrule
            \midrule
            & ERM & 29.24$\pm$0.81 & 39.67$\pm$4.17 & 25.85$\pm$3.51 \\
            & GTrans & 29.24$\pm$0.85 & 39.65$\pm$3.93 & 25.62$\pm$3.63 \\
            \rowcolor{gray!30}
            \multirow{-3}*{\rotatebox{90}{GCN}}\cellcolor{white} &\textbf{GrapHoST } & \color{purple}\textbf{32.57$\pm$1.26} & \color{purple}\textbf{40.79$\pm$3.82} & \color{purple}\textbf{26.32$\pm$3.04} \\
             
            \midrule
            & ERM & 28.96$\pm$1.30 & 36.36$\pm$2.24 & 24.51$\pm$2.16 \\
            & GTrans & 28.95$\pm$1.30 & 36.47$\pm$2.29 & 24.50$\pm$2.14 \\
            \rowcolor{gray!30}
            \multirow{-3}*{\rotatebox{90}{GAT}}\cellcolor{white} & \textbf{GrapHoST } & \color{purple}\textbf{32.43$\pm$2.11} & \color{purple}\textbf{37.49$\pm$2.89} & \color{purple}\textbf{25.11$\pm$1.88} \\

             \midrule
            & ERM & 28.45$\pm$3.58 & 38.27$\pm$0.89 & 26.21$\pm$1.35 \\
            & GTrans & 28.40$\pm$3.55 & 38.33$\pm$1.12 & 26.22$\pm$1.28 \\
            \rowcolor{gray!30}
            \multirow{-3}*{\rotatebox{90}{SAGE}}\cellcolor{white} &\textbf{GrapHoST } & \color{purple}\textbf{28.49$\pm$3.59} & \color{purple}\textbf{38.73$\pm$1.26} & \color{purple}\textbf{26.30$\pm$1.24} \\
             
            \midrule
            & ERM & 33.39$\pm$0.66 & 45.26$\pm$0.75 & 30.19$\pm$0.51 \\
            & GTrans & 33.38$\pm$0.67 & 45.22$\pm$1.20 & 30.35$\pm$0.32 \\
            \rowcolor{gray!30}
            \multirow{-3}*{\rotatebox{90}{GPR}}\cellcolor{white} & \textbf{GrapHoST } & \color{purple}\textbf{35.53$\pm$0.64} & \color{purple}\textbf{45.46$\pm$0.73} & \color{purple}\textbf{31.92$\pm$0.80} \\

            \midrule
            & ERM & 33.75$\pm$0.98 & 48.57$\pm$1.34 & 29.91$\pm$2.07 \\
            & GTrans & 33.86$\pm$1.27 & 48.46$\pm$1.55 & 28.33$\pm$3.10 \\
            \rowcolor{gray!30}
            \multirow{-3}*{\rotatebox{90}{H2GCN}}\cellcolor{white} & \textbf{GrapHoST } & \color{purple}\textbf{34.11$\pm$0.81} & \color{purple}\textbf{51.37$\pm$1.68} & \color{purple}\textbf{29.94$\pm$2.48} \\
            
            \bottomrule
        \end{tabular}
    }\label{tab:heter}
\end{table}

\subsection{Effectiveness on Filtered Heterophilic Datasets with Different Splits}
\label{sec:filter}
Following the split settings from GTrans and EERM, we evaluate heterophilic datasets Actor, Chameleon, and Squirrel under a node attribute shift, comparing them with homophilic datasets such as Cora and Amazon. To address potential node redundancy, we further test both original and filtered versions of Chameleon and Squirrel using the splits from~\cite{Geom-GCN} and the dataset filtering process from~\cite{heter}. While prior work uses a GNN hidden dimension of 512, we adopt 32 for consistency with GTrans. As shown in Table~\ref{tab:filtered-heterophilic}, performance on filtered Squirrel remains stable, whereas filtered Chameleon decreases notably due to data leakage~\cite{heter}. \textbf{Nonetheless, GrapHoST consistently outperforms GTrans and base GNNs, demonstrating its robustness across settings.}

\begin{table}[!t]
    \centering
    \caption{Performance on filtered ($^*$) and unfiltered heterophilic datasets (accuracy \%) with original split setting.}
    \resizebox{\linewidth}{!}{
        \begin{tabular}{l|cccc}
            \toprule
            \textbf{Method} & \textbf{Chameleon} & \textbf{Chameleon$^*$} & \textbf{Squirrel} & \textbf{Squirrel$^*$} \\
            \midrule
            \midrule
            \textbf{GCN} & \color{teal}\underline{39.67$\pm$4.17} & \color{teal}\underline{28.82$\pm$6.64} & \color{teal}\underline{25.85$\pm$3.51} & \color{teal}\underline{25.73$\pm$3.59} \\
            \textbf{+ GTrans} & 39.65$\pm$3.93 & 27.79$\pm$5.58 & 25.62$\pm$3.63 & 24.89$\pm$3.67 \\
            \textbf{+ GrapHoST} & \color{purple}\textbf{40.72$\pm$3.85} & \color{purple}\textbf{30.06$\pm$5.90} & \color{purple}\textbf{26.32$\pm$3.04} & \color{purple}\textbf{26.04$\pm$3.32} \\
            \midrule
            \textbf{GAT} & 36.36$\pm$2.24 & 27.75$\pm$3.88 & \color{teal}\underline{24.51$\pm$2.16} & \color{teal}\underline{24.69$\pm$2.44} \\
            \textbf{+ GTrans} & \color{teal}\underline{36.47$\pm$2.29} & \color{teal}\underline{27.82$\pm$4.12} & 24.50$\pm$2.14 & 24.60$\pm$2.25 \\
            \textbf{+ GrapHoST} & \color{purple}\textbf{37.44$\pm$3.51} & \color{purple}\textbf{28.86$\pm$3.35} & \color{purple}\textbf{25.09$\pm$1.93} & \color{purple}\textbf{25.11$\pm$2.03} \\
            \midrule
            \textbf{H2GCN} & 40.13$\pm$2.89 & \color{teal}\underline{22.57$\pm$4.31} & 27.08$\pm$1.96 & 29.82$\pm$2.06 \\
            \textbf{+ GTrans} & \color{teal}\underline{40.24$\pm$3.18} & 21.96$\pm$4.72 & \color{teal}\underline{27.74$\pm$2.40} & \color{teal}\underline{29.89$\pm$4.13} \\
            \textbf{+ GrapHoST} & \color{purple}\textbf{42.24$\pm$3.97} & \color{purple}\textbf{24.21$\pm$4.05} & \color{purple}\textbf{28.93$\pm$2.23} & \color{purple}\textbf{31.26$\pm$3.37} \\
            \bottomrule
        \end{tabular}
    }
    \label{tab:filtered-heterophilic}
\end{table}

\subsection{Test-time Graph Homophily Shift Analysis}
\label{sec:homo-shift}
For graphs with cross-domain or temporal evolution issues, a clear shift in edge homophily degree is observed across the graphs after the split, as shown in Figure~\ref{fig:homo_shift}. For example, the OGB-Arxiv dataset, a paper citation network collected from 2014 to 2020, exhibits a natural temporal shift. The objective is to classify papers, with training papers selected from those published before 2011, validation papers from 2011 to 2014, and test papers from the periods 2014-2016, 2016-2018, and 2018-2020. A clear evolution of homophily patterns is observed over time, and the homophily degree gradually decreases from 0.73 in the training dataset to 0.66 in the final test dataset, which can be attributed to the emergence of more diverse paper types within different classes and their interconnections as time progresses. Furthermore, for the Elliptic dataset, when the dataset is split based on the temporal dimension, a clear fluctuating homophily pattern is also observed across the graphs. Combining such observations with the results in Table 2, \textbf{GrapHoST achieves superior performance across these datasets, underscoring its robustness to shifts in homophily patterns.}

For the Facebook-100 dataset, 100 snapshots of Facebook friendship networks from 2005 are included, each representing users from specific American universities. A total of fourteen networks are adopted: John Hopkins, Caltech, Amherst, Bingham, Duke, Princeton, WashU, Brandeis, Carnegie, Cornell, Yale, Penn, Brown, and Texas. Among them, Penn, Brown, and Texas are used for testing, Cornell and Yale for validation, and the remaining graphs are split into three different combinations for training. 

Previous work~\cite{EERM} has analysed the significant changes in node number, density, and degree distribution across training, validation, and test graphs. In contrast, in our work, the homophily-based properties are the main focus. According to Figure~\ref{fig:homo_shift}, the average edge homophily degree is calculated and averaged across all the training, validation, and three groups of test graphs following the split from GTrans~\cite{GTRANS}. While the training and validation graphs exhibit similar edge homophily patterns, the test graph homophily patterns shift significantly: the first two groups of test graphs show notably lower edge homophily degrees, while the last group of test graphs showcase higher edge homophily degrees compared to the training and validation graphs. 

Furthermore, according to the dataset statistic table, it is clear that the average edge homophily degree of the FB-100 dataset is significantly lower than other homophilic datasets, indicating that FB-100 contains a mixture of homophilic and heterophilic patterns. As shown in the results in Table 2, existing methods such as GTrans and EERM have struggled to achieve improvements on the FB-100 dataset. In contrast, \textbf{GrapHoST achieves substantial performance gains and outperforms all baseline methods on FB-100, highlighting its effectiveness on graphs exhibiting a complex mixture of homophily and heterophily patterns.}

\section{More Details on Related Work}
\label{app:related}

\subsection{Test-time Graph Data Quality Issues}
\label{app:quality}
Test-time graph data quality issues have presented a significant challenge to the deployment of GNNs~\cite{EERM}. For example, GNNs often exhibit sub-optimal performance when there is a misalignment between the training and test graph distributions~\cite{EERM,zhu2021shift,liu2022confidence,chen2022invariance,buffelli2022sizeshiftreg,alsa,gold,tntood,puma,cat,caselink,casegnn}. Moreover, graph structures from different sources can be disrupted by human errors, outliers, and structure attacks at test-time, leading to degradation in the performance of the pre-trained GNNs~\cite{zugner2018adversarial,li2021adversarial}.

\subsection{Test-time Operations on Graph}
\subsubsection{Test-time Training of Graph Neural Networks}
\label{app:operation_1}
To address various test-time data quality issues, test-time training was proposed to update the model at test time. In the graph domain, GAPGC~\cite{GraphTTA} improves the adaptation of GNNs using adversarial graph pseudo-group contrast methods. GT3~\cite{GT3} adopt self-supervised learning to tailor GNNs to test graphs. Similarly, HomoTTT~\cite{HomoTTT} develops a parameter-free contrastive loss to retrain GNNs on the test graphs. After that, LLMTTT~\cite{LLMTTT} retrains GNN on test-attributed graphs annotated by large language models. Moreover, LEBED~\cite{lebed} develops a learning behaviour discrepancy score by retraining additional GNNs to evaluate the original GNN performance during test-time. AdaRC~\cite{Adarc} adjusts the hop-aggregation parameters in GNNs during test time to mitigate the impact of distribution shift. \textbf{In general, all those methods put their efforts into solving the test graph data quality issues from the model perspective} by improving the model architecture or retraining the model parameters without altering the input graph data. In contrast, the proposed GrapHoST method targets the test-time graph data quality issues \textbf{from a data point of view by directly improving the test graph structure.}

\subsubsection{Test-time Transformation of Graph}
\label{app:operation_2}
A few studies address test graph data quality issues by directly transforming the graphs during test-time~\cite{GTRANS, GP}. GTrans~\cite{GTRANS} proposes a pioneering test-time graph transformation paradigm that enhances graph features and structure by performing contrastive learning on test graphs with data quality issues. GraphPatcher~\cite{GP} trains an additional model to patch nodes in the test graphs. Such test-time methods demonstrate a promising potential to enhance the test-time performance of the pre-trained GNN by improving the quality of test data, without the need for retraining or modifying the architecture of the pre-trained GNN. \textbf{Unlike previous work, GrapHoST leverages homophily-based properties within the graph structure -- an aspect neglected in prior studies -- to further improve test data quality and enhance the performance of the fixed GNNs.}

\subsection{Homophily and Heterophily on Graph}
\label{app:homophily}
As important properties of graph structure, previous work reveals that homophily and heterophily can significantly affect the GNN training procedure, thus affecting model performance in various domains~\cite{Geom-GCN, zhu2020beyond, GOAL, bo2021beyond, GPR, yan2022two, luan2021heterophily, li2022finding, ACM, ma2021homophily, DCGC}. Compared to existing methods, \textbf{GrapHoST seeks to improve GNN performance by taking advantage of homophily-based properties via graph structural transformation at test-time without modifying the model architecture or the training phase, which is a completely different and under-explored field.}

\section{Further Discussion on GrapHoST}
\label{app:discuss}
\subsection{Relations to Training-time Methods}
\label{app:gsl}
Homophily-based properties are also crucial during GNN training, particularly in the domain of graph structure learning (GSL). For example, OpenGSL~\cite{Opengsl} offers a comprehensive investigation of the relationship between homophily and GNN effectiveness during training time. However, GrapHoST is specifically designed for test-time graph structural transformation and cannot be directly applied during training-time, as ground truth labels are available. For example, if GrapHoST were applied to homophilic graphs during training, it would remove all heterophilic edges, resulting in a graph where nodes from different classes are entirely disconnected. This would transform the node classification task into a trivial problem, especially if label propagation is employed.

\subsection{Relations to Link Prediction and Clustering}
\label{app:link}
Link prediction and graph clustering are closely related research areas that focus on predicting node linkages or communities. For example, SEAL~\cite{SEAL} is designed to predict the existence of links by leveraging sub-graph patterns. Similarly, DMoN~\cite{cluster} introduces an unsupervised pooling method to identify clustering structures in real-world graphs. In contrast, GrapHoST addresses a fundamentally different problem: \textbf{transforming the test graph structure by classifying edges based on homophily-based properties to enhance the test-time performance of the pre-trained GNNs.}

\subsection{Discussion on Node Attribute Shift}
\label{app:node_attribute}

To address the challenges posed by node attribute shifts in GNN test-time performance, the test graph structure can be transformed into a more suitable configuration that facilitates effective message passing in GNNs. In homophilic test graphs with node feature shift issues, ineffective message passing often arises under the original structure. Applying homophily-based structural transformation during test-time strengthens the connectivity between nodes and their neighbours that share the same label, thereby enhancing the aggregation of valuable messages from neighbours with the same label to nodes with perturbed attributes. Similarly, in heterophilic graphs with node attribute shifts, the homophily-based structural transformation reinforces heterophilic edge connectivity patterns in the test graph, enabling effective message passing in GNNs that are well-trained on such heterophilic graphs. Grounded in Theorems 1, Theorems 2 and the visualisation of the GNN output embedding of GrapHoST on graphs with node attribute shift in the main paper, homophily-based graph transformations facilitate the generation of more class-separable node embeddings, thereby mitigating the impact of node attribute shifts and improving the performance of fixed pre-trained GNN at test-time.

\subsection{Discussion on Test-time Paradigms}
\label{app:test-time}
In the broad area of test-time data transformation for general machine learning problems, a transformation policy can be learned jointly with the base classifier by training data, as supported by existing methods across various domains, including computer vision~\cite{tt0, tt1} and graph learning~\cite{GP}. For example, an auxiliary loss predictor is trained with labelled data for better test-time augmentations with the lowest predicted loss~\cite{tt0}. Moreover, GraphPatcher trained an additional node generator model to patch nodes into the test graphs~\cite{GP}. In line with existing research~\cite{tt0, tt1, GP}, GrapHoST chooses to train an auxiliary homophily predictor, targeting at exploiting homophily-based properties in the graph structure to improve GNN test-time performance without access to ground-truth test labels. Thus, GrapHoST can be safely categorised as a test-time graph transformation method.

\end{document}